\pdfoutput=1

\documentclass[11pt]{article}

\usepackage[preprint]{acl}

\usepackage[show]{chato-notes}
\usepackage{times}
\usepackage{latexsym}
\usepackage{soul}
\usepackage{makecell}
\usepackage{pifont}

\usepackage[T1]{fontenc}

\usepackage[utf8]{inputenc}

\usepackage{microtype}

\usepackage{inconsolata}

\usepackage{graphicx}
\usepackage{numprint}
\usepackage{booktabs}
\usepackage{float}

\usepackage{siunitx}
\usepackage{amsmath}
\usepackage{pgfplots}
\usepackage{pgfplotstable}
\usepackage{tikz}
\usetikzlibrary{positioning, shapes, plotmarks}
\usepgfplotslibrary{polar}
\usepgfplotslibrary{statistics}
\usepackage{listings}

\newcommand{\llamatwo}{Llama 2-7b}
\newcommand{\opt}{OPT-6.7b}

\usepackage{multirow}
\usepackage{csquotes}
\usepackage{enumitem}
\usepackage{soul}

\usepackage{fancyhdr} 
\makeatletter
\def\ps@firstpagestyle{\ps@fancy}
\makeatother

\title{Are the Hidden States Hiding Something? Testing the Limits of Factuality-Encoding Capabilities in LLMs
\thanks{
This is the authors’ version of the work. The final, published version will appear in the \textit{Proceedings of the 63rd Annual Meeting of the Association for Computational Linguistics (ACL `25)}.\\ 

This work is licensed under a \href{https://creativecommons.org/licenses/by/4.0/}{Creative Commons Attribution 4.0 International License (CC BY 4.0)}.\\

Please cite the official published version when available.}
}

\author{
  \normalsize
  \textbf{Giovanni Servedio\textsuperscript{1,2\dag}} \hspace{1em}
  \textbf{Alessandro De Bellis\textsuperscript{1\dag}} \hspace{1em}
  \textbf{Dario Di Palma\textsuperscript{1}} \\
  \normalsize
  \textbf{Vito Walter Anelli\textsuperscript{1}} \hspace{1em}
  \textbf{Tommaso Di Noia\textsuperscript{1}} \\
  \normalsize
  \textsuperscript{1}Politecnico di Bari, Italy \hspace{1em}
  \textsuperscript{2}Sapienza University of Rome, Italy \\
  \normalsize
  \href{mailto:giovanni.servedio@poliba.it, alessandro.debellis@poliba.it, dario.dipalma@poliba.it}{name.surname@poliba.it}
}

\hypersetup{
    colorlinks=true,
    linkcolor=black,
    urlcolor=black,
    citecolor=black
}

\begin{document}
\maketitle

\renewcommand{\thefootnote}{\dag}  
\footnotetext{Corresponding authors.}
\renewcommand{\thefootnote}{\arabic{footnote}} 

\begin{abstract}
Factual hallucinations are a major challenge for Large Language Models (LLMs). They undermine reliability and user trust by generating inaccurate or fabricated content.
Recent studies suggest that when generating false statements, the internal states of LLMs encode information about truthfulness.
However, these studies often rely on synthetic datasets that lack realism, which limits generalization when evaluating the factual accuracy of text generated by the model itself.
In this paper, we challenge the findings of previous work by investigating truthfulness encoding capabilities, leading to the generation of a more realistic and challenging dataset. Specifically, we extend previous work by introducing: (1) a strategy for sampling plausible true-false factoid sentences from tabular data and (2) a procedure for generating realistic, LLM-dependent true-false datasets from Question Answering collections. Our analysis of two open-source LLMs reveals that while the findings from previous studies are partially validated, generalization to LLM-generated datasets remains challenging. This study provides a foundation for future research on factuality in LLMs and offers practical guidelines for more effective evaluation. Code is provided at our \href{https://github.com/sisinflab/HidingInTheHiddenStates}{ \texttt{GitHub Repository}}.

\end{abstract}
\section{Introduction}
In the last few years, Large Language Models (LLMs) have shown outstanding abilities in natural language processing tasks and beyond~\cite{biancofiore2025conversational, DBLP:conf/recsys/Palma23}.
Nevertheless, factual hallucinations~\cite{zhang2023siren} represent a significant obstacle, limiting their reliability and hindering their safe deployment in real-world applications~\cite{di2025llms} such as healthcare~\cite{medical}, education~\cite{UpadhyayGC23}, legal advice~\cite{Dahl2024}, and language understanding~\cite{DBLP:conf/semweb/BellisANS24, DBLP:conf/cikm/AnelliBBNS22}. Hallucinations occur when an LLM generates content that is syntactically coherent but factually inaccurate, decreasing trust in AI systems~\cite{Huang2024-vq}. 
Recent research suggests that LLMs may encode internal representations of factuality in their hidden states, indicating an awareness of whether a generated statement is true or false~\cite{chen2024inside}. These efforts led to the development of approaches to evaluate the factual accuracy of the LLM outputs given their internal representations (\textit{factuality "self-evaluation"}).
Self-evaluation can be used to identify gaps in the knowledge of an LLM, improving truthfulness and transparency through abstention mechanisms \cite{feng-etal-2024-dont}, fact verification \cite{wadden-etal-2020-fact}, and self-correction \cite{ji-etal-2023-towards}.
\citet{azaria2023llmlying} suggest that LLMs have \enquote{some internal notion as to whether a sentence is true or false, as this information is required for generating (or predicting) following tokens.} Based on this assumption, they propose a neural classifier to discern factual from non-factual statements based on hidden layer activations.
However, the datasets used to evaluate the probe present limitations since they contain trivially incorrect statements (e.g., "The zebra uses flying for locomotion") that easily fail to align with the generative patterns of LLMs. Additionally, the false statements are generated using random substitutions of the true terms with little regard for the plausibility of negative samples. This misalignment not only weakens the generalizability of results but also raises concerns about the applicability of these models to real-world scenarios where false statements may be subtle or nuanced.
This study addresses these gaps by generating more plausible datasets (see Figure~\ref{fig:radar_comparison}) to explore LLM factuality encoding and evaluating refined models.
The primary contributions are:
\begin{enumerate}[label=\arabic*., itemsep=1pt, parsep=1pt, topsep=1pt, partopsep=1pt, align=left, leftmargin=*]
    \item We \textbf{reproduce} the methodology of~\citet{azaria2023llmlying} to ensure transparency.
    \item We \textbf{propose two strategies to generate realistic datasets} and discuss how well the original and newly developed models generalize.
\end{enumerate}
Specifically, we extend their work with two novel dataset creation strategies and design a strategy that better fits the factuality self-evaluation task, introducing:
\begin{itemize}[itemsep=1pt, parsep=1pt, topsep=1pt, partopsep=1pt, align=left, leftmargin=*]
    \item A \textbf{perplexity-based negative sampling strategy} that enhances the original generation mechanism and leverages the LLM token distribution. 
    \item A \textbf{novel strategy to sample realistic LLM-generated facts}, leveraging Question Answering datasets to elicit responses from the LLM. 
\end{itemize}
Through this analysis, we lay the groundwork for more robust factuality assessments and offer practical guidelines for enhancing the reliability of LLMs in diverse applications.

\section{Reproduction of Prior Work: Settings} \label{sec:replication}
In their study, \citet{azaria2023llmlying} investigate whether LLMs internally represent the factuality of sentences. This section summarizes the dataset generation approach they employed and the specific probing architecture used in their study.


\subsection{Dataset Generation Strategy}\label{sec:dataset}

To explore whether LLMs internally represent the factuality of statements, the authors constructed a `True-False' dataset of facts labeled as either True or False, covering six disjoint topics: \textit{Cities}, \textit{Inventions}, \textit{Chemical Elements}, \textit{Animals}, \textit{Companies}, and \textit{Scientific Facts}. To generate the dataset, for the first five topics, i.e. with the exception of \textit{Scientific Facts}, the authors selected \textbf{true statements} from reliable sources (see Table~\ref{tab:reliable_source} in the appendix) and produced \textbf{false statements}, replacing part of a true statement (e.g. \textit{“Hydrogen has an atomic number of 1”}) with randomly sampled incorrect information (\textit{“Hydrogen has an atomic number of 34”}). Meanwhile, for \textit{Scientific Facts}, they employed ChatGPT (13 Feb 2023) as a generator of true and false sentences, and two human annotators manually verified their correctness. The authors publicly release the dataset, which we refer to as the \textit{“True-False dataset”}.
Furthermore, the authors constructed a second dataset using the \opt{} model, which we refer to as the `\textbf{OPT-Generated Dataset}'. To create this dataset, the model was prompted with a true statement absent from the True-False dataset and then used to generate a subsequent sentence. The responses were manually fact-checked and annotated by three independent human judges. Non-factual responses were filtered out, resulting in a final set of 245 statements.

\subsection{Internal States Analysis via SAPLMA}
To investigate whether LLMs internally represent the factuality of statements, \citet{azaria2023llmlying} developed a probe (\textbf{S}tatement \textbf{A}ccuracy \textbf{P}rediction based on \textbf{L}anguage \textbf{M}odel \textbf{A}ctivations) that predicts the factual accuracy of a statement by analyzing the hidden layer activations of an LLM. 
SAPLMA is a feedforward neural network designed to classify statements as true or false. It consists of three hidden layers (256, 128, 64) and a sigmoid output activation. The model is trained using the Adam optimizer for five epochs without hyperparameter tuning.
The authors studied two LLMs, namely \opt{}~\cite{zhang2022opt} and \llamatwo{}~\cite{DBLP:journals/corr/abs-2307-09288}, both consisting of 32 layers. To identify which layers best capture factuality, they trained five SAPLMA models, forwarding each statement in the True-False Dataset as input to the LLMs and extracting the corresponding activation values from the 32nd (last layer), 28th, 24th, 20th, and 16th 
layers. These activations serve as input for training the classifiers.
To ensure generalizability, i.e., making SAPLMA independent of specific topics, the authors adopted a cross-validation strategy using a leave-one-topic-out approach to train the classifier on five topics and test the probe on the held-out topic.

\subsection{Reproducibility Settings}
In this section, we provide details on the datasets and experimental settings for reproducing the work of \citet{azaria2023llmlying}. Our goal is to answer the Research Question (\textbf{RQ0}): \textit{Can we reproduce the results reported by \citet{azaria2023llmlying}?"}
Although the code is not publicly accessible, the authors made it available upon request. 

\vspace{0.2em}
\noindent \textbf{Dataset generation.} The code provided by \citet{azaria2023llmlying} contains all the necessary material to recreate the entire dataset generation process. However, their template-matching code is influenced by randomness in the generation of false statements, and a random seed is not set. Due to this non-deterministic behavior, recreating their dataset using the original code was unfeasible. However, the authors released their dataset, allowing us to reproduce their exact dataset settings.

\begin{table*}[ht!]
\centering
\resizebox{\textwidth}{!}{%
\begin{tabular}{llcccccccccccccc}
\toprule
\multicolumn{2}{c}{\multirow{2}{*}{\textbf{Layer}}} & \multicolumn{2}{c}{\textbf{Cities}} & \multicolumn{2}{c}{\textbf{Inventions}} & \multicolumn{2}{c}{\textbf{Elements}} & \multicolumn{2}{c}{\textbf{Animals}} & \multicolumn{2}{c}{\textbf{Companies}} & \multicolumn{2}{c}{\textbf{Facts}} & \multicolumn{2}{c}{\textbf{Average}} \\ 
\cmidrule(lr){3-4} \cmidrule(lr){5-6} \cmidrule(lr){7-8} \cmidrule(lr){9-10} \cmidrule(lr){11-12} \cmidrule(lr){13-14} \cmidrule(lr){15-16} 
& & Llama2 & OPT-6.7b & Llama2 & OPT-6.7b & Llama2 & OPT-6.7b & Llama2 & OPT-6.7b & Llama2 & OPT-6.7b & Llama2 & OPT-6.7b & Llama2 & OPT-6.7b \\ 
\midrule
\multirow{2}{*}{\textbf{Last}} & Orig. & 0.7574 & 0.7796 & 0.6735 & 0.5696 & 0.6814 & 0.5760 & 0.7338 & 0.6022 & 0.6736 & 0.6925 & 0.7444 & 0.6498 & 0.7107 & 0.6449 \\ 
& Repr. & 0.7939 & 0.7836 & 0.7470 & 0.5603 & 0.7057 & 0.5656 & 0.7133 & 0.5984 & 0.6463 & 0.6900 & 0.7894 & 0.6640 & 0.7326 & 0.6437 \\ 
\midrule
\multirow{2}{*}{\textbf{28}} & Orig. & 0.8146 & 0.7732 & 0.7207 & 0.5761 & 0.6767 & 0.5907 & 0.7249 & 0.5777 & 0.6894 & 0.7247 & 0.7662 & 0.6618 & 0.7321 & 0.6507 \\ 
& Repr. & 0.8261 & 0.8014 & 0.7221 & 0.5938 & 0.6746 & 0.5931 & 0.7046 & 0.5945 & 0.6860 & 0.7252 & 0.7976 & 0.6639 & 0.7351 & 0.6620 \\ 
\midrule
\multirow{2}{*}{\textbf{24}} & Orig. & 0.8722 & 0.7963 & 0.7816 & 0.6712 & 0.6849 & 0.6211 & 0.7394 & 0.5800 & 0.7094 & 0.7758 & 0.7858 & 0.6868 & 0.7622 & 0.6886 \\ 
& Repr. & 0.8619 & 0.8043 & 0.7737 & 0.6604 & 0.6789 & 0.6172 & 0.7415 & 0.6095 & 0.7049 & 0.7844 & 0.7910 & 0.6804 & 0.7586 & 0.6927 \\ 
\midrule
\multirow{2}{*}{\textbf{20}} & Orig. & 0.8820 & \textbf{0.8125} & 0.8459 & \textbf{0.7268} & 0.6950 & 0.6197 & 0.7758 & 0.6058 & 0.8319 & 0.8122 & 0.8053 & \textbf{0.6819} & 0.8060 & 0.7098 \\ 
& Repr. & 0.8672 & 0.8118 & 0.8584 & 0.7222 & 0.6761 & \textbf{0.6218} & 0.7736 & \textbf{0.6208} & 0.8254 & \textbf{0.8160} & 0.8065 & 0.6734 & 0.8012 & \textbf{0.7110} \\ 
\midrule
\multirow{2}{*}{\textbf{16}} & Orig. & \textbf{0.9223} & 0.7435 & \textbf{0.8938} & 0.6400 & 0.6939 & 0.5645 & 0.7774 & 0.5800 & 0.8658 & 0.7570 & \textbf{0.8254} & 0.6237 & \textbf{0.8298} & 0.6515 \\ 
& Repr. & 0.9174 & 0.7554 & 0.8847 & 0.6403 & \textbf{0.7005} & 0.5732 & \textbf{0.7883} & 0.5693 & \textbf{0.8672} & 0.7760 & 0.8104 & 0.6340 & 0.8281 & 0.6580 \\ 
\bottomrule
\end{tabular}%
}
\caption{Replicated SAPLMA performance on the True-False dataset across the selected layers. The results labeled as `Orig.' are taken from the original work, while those labeled as `Repr.' are the replicated results from this study.}
\label{tab:sampla_reprod-results}
\vspace{-0.5cm}
\end{table*}

\noindent \textbf{SAPLMA reproducibility.} To reproduce the results of the original study, we trained 20 SAPLMA probes for each of the following layers: the 32nd, 28th, 24th, 20th, and 16th, over 5 epochs, resulting in a total of 100 probes. We employed \llamatwo{} and \opt{}, both using half-precision (16-bit float) parameters, with a default temperature of 0.8 for \llamatwo{} and 1.0 for \opt{}. The hardware used for the experiments was an Intel(R) Core(TM) i7-5820K paired with an NVIDIA RTX 3090 graphics card.
The authors do not specify from which token they extract the associated hidden state. 
However, code inspection led to the identification of the last token as the target state.

\begin{table}[ht!]
\centering
\resizebox{\columnwidth}{!}{%
\begin{tabular}{llccccccc}
\toprule
\multicolumn{2}{c}{\textbf{Model}} & \textbf{Cities} & \textbf{Inventions} & \textbf{Elements} & \textbf{Animals} & \textbf{Companies} & \textbf{Facts} & \textbf{Average} \\ 
\cmidrule(lr){1-2} \cmidrule(lr){3-9} 
\multirow{2}{*}{\textbf{BERT}}       & Orig. & 0.5357 & 0.5537 & 0.5645 & 0.5228 & 0.5533 & 0.5302 & 0.5434 \\ 
                                     & Repr. & 0.5257 & 0.5611 & 0.5435 & 0.5603 & 0.5302 & 0.5361 & 0.5428 \\ 
\midrule
\multirow{2}{*}{\textbf{3-shot}}     & Orig. & 0.5410 & 0.4799 & 0.5685 & 0.5650 & 0.5538 & 0.5164 & 0.5374 \\ 
                                     & Repr. & 0.5416 & 0.4800 & 0.5685 & 0.5652 & 0.5539 & 0.5115 & 0.5368 \\ 
\midrule
\multirow{2}{*}{\textbf{5-shot}}     & Orig. & 0.5416 & 0.4799 & 0.5676 & 0.5643 & 0.5540 & 0.5148 & 0.5370 \\ 
                                     & Repr. & 0.5416 & 0.4800 & 0.5676 & 0.5643 & 0.5540 & 0.5082 & 0.5359 \\ 
\midrule
\multirow{2}{*}{\textbf{It-is-true}} & Orig. & 0.5230 & 0.5068 & 0.5688 & 0.4851 & 0.6883 & 0.5840 & 0.5593 \\ 
                                     & Repr. & 0.5233 & 0.5046 & 0.5688 & 0.4831 & 0.6875 & 0.5856 & 0.5588 \\ 
\bottomrule
\end{tabular}%
}
\caption{Replicated baselines performance on the True-False dataset. The results labeled as ‘Orig.’ are taken from the original work, while those labeled as ‘Repr.’ are the replicated results from this study.}
\label{tab:baselines-reproduced}
\end{table}
\begin{table}[ht!]
\centering
\resizebox{\columnwidth}{!}{%
\begin{tabular}{llcccc}
\toprule
\textbf{Layer} & & \textbf{Accuracy} & \textbf{AUC} & \textbf{Accuracy with} & \textbf{Average Threshold} \\ 
& & & & \textbf{Optimal Threshold} & \\
\midrule
\multirow{2}{*}{\textbf{Last-layer}}   & Orig. & 0.6187 & 0.7587 & 0.7052 & 0.8687 \\ 
                                       & Repr. & 0.6406 & 0.7720 & 0.7264 & 0.8910 \\ 
\midrule
\multirow{2}{*}{\textbf{28th-layer}}   & Orig. & 0.6362 & 0.7614 & 0.7134 & 0.8838 \\ 
                                       & Repr. & 0.6410 & 0.7686 & 0.7203 & 0.8276 \\ 
\midrule
\multirow{2}{*}{\textbf{24th-layer}}   & Orig. & 0.6134 & 0.7435 & 0.6988 & 0.8801 \\ 
                                       & Repr. & 0.6206 & 0.7496 & 0.6973 & 0.8500 \\ 
\midrule
\multirow{2}{*}{\textbf{20th-layer}}   & Orig. & 0.6029 & 0.7182 & 0.6587 & 0.9063 \\ 
                                       & Repr. & 0.5965 & 0.7183 & 0.6669 & 0.8868 \\ 
\midrule
\multirow{2}{*}{\textbf{Middle-layer}} & Orig. & 0.5566 & 0.6610 & 0.6500 & 0.8123 \\ 
                                       & Repr. & 0.5579 & 0.6760 & 0.6468 & 0.7948 \\ 
\midrule
\multirow{2}{*}{\textbf{BERT}}         & Orig. & 0.5115 & 0.5989 & 0.5705 & 0.9403 \\ 
                                       & Repr. & 0.5522 & 0.6092 & 0.5689 & 0.7939    \\ 
\midrule
\multirow{2}{*}{\textbf{3-shot}}       & Orig. & 0.5041 & 0.4845 & -      & -      \\ 
                                       & Repr. & 0.5041      & 0.4845      & -      & -      \\ 
\midrule
\multirow{2}{*}{\textbf{5-shot}}       & Orig. & 0.5125 & 0.4822 & -      & -      \\ 
                                       & Repr. & 0.5125      & 0.4822      & -      & -      \\ 
\bottomrule
\end{tabular}%
}
\caption{Reproduced SAPLMA performance on the \textbf{OPT-Generated Dataset} (Section \ref{sec:dataset}). The results labeled as `Orig.' are taken from the original work, while those labeled as `Repr.' are the reproduced results.}
\label{tab:reproduced-generated-dataset}
\end{table}
\section{Experimental Reproducibility Results}
To answer \textbf{RQ0}, we report the results for the reproduction of the experiments in Tables~\ref{tab:sampla_reprod-results}, \ref{tab:baselines-reproduced} and \ref{tab:reproduced-generated-dataset}. The results labeled with `Orig.' are retrieved from the original work, while the ones achieved in the reproducibility study are labeled with `Repr.'.

\subsection{Reproduction of SAPLMA Results on the True-False Dataset}
Table~\ref{tab:sampla_reprod-results} reports SAPLMA's performance on the True-False dataset across different layers of \llamatwo{} and \opt{} for six categories: Cities, Inventions, Elements, Animals, Companies, and Facts.
Results indicate that \llamatwo{} consistently outperforms \opt{} across all layers and categories. Middle layers (16, 20, 24) achieve the highest performance, while accuracy declines toward the final layer. The reproduced results closely align with the original findings, with minor deviations observed across specific categories and layers.
Moreover, although \opt{} shows greater variability, the ranking of layers remains consistent with the original work. 
Both experiments confirm that factuality information is more effectively encoded in the middle layers (16–24) than in the final layer.
Additionally, we reproduce the baseline used by the authors to compare SAPLMA. Specifically, their baseline includes a trained SAPLMA on BERT activations, a few-shot approach where the LLM is prompted with a sentence and asked to label it as `True' or `False,' and an `It-is-true' test. In this test, the LLMs were asked: \textit{Is it true} that X? and \textit{Is it false that X?}, where X is a dataset sample. A response was considered correct if the model assigned a higher probability to the `True' token. Table~\ref{tab:baselines-reproduced} summarizes the baseline performance.
BERT achieves the highest average performance among non-prompted methods, demonstrating its effectiveness in factual classification, and the results indicate high reproducibility.
The It-is-true baseline yields the highest average performance, particularly in the `Companies' topic. The results demonstrate consistent reproducibility.

In general, the reproduction on the True-False Dataset is considered successful based on the following observations:
(i) the overall performance trends remain consistent, with \llamatwo{} outperforming \opt{};
(ii) observed deviations are minor and do not indicate fundamental inconsistencies;
(iii) the relative ranking of layers remains unchanged, reinforcing previous findings; and
(iv) baseline methods retain their rankings, confirming the validity of the original results.

\subsection{Reproduction of the Results on the OPT-Generated Dataset}\label{sec:3-generated}
Table~\ref{tab:reproduced-generated-dataset} presents the results of the reproduced experiments on the OPT-Generated Dataset. Performance is evaluated using Accuracy, AUC, Accuracy with an Optimal Threshold (selected by estimating it from a held-out validation set), and the Average Optimal Threshold.
The final layers, specifically the 28th and final layers, outperform the middle and lower layers in terms of Accuracy and AUC. The reproduced results closely align with the original findings, exhibiting only minor variations. 
Accuracy with anoptimal threshold consistently exceeds raw accuracy, suggesting that tuning the decision boundary improves 
performance. 

Regarding baselines, BERT exhibits lower accuracy compared to the LLMs' last layers, with slight improvements over the original results. 
Notably, the 3- and 5-shot prompting results were identical between the original and reproduced experiments.

\vspace{0.2em}
\textbf{RQ0}: \textit{Can we reproduce the results reported by \citet{azaria2023llmlying}?} 

\textbf{This reproducibility study demonstrates a high degree of alignment with the original results}, confirming the validity of previous findings. It shows that the ranking and trends remain unchanged, reinforcing the robustness of the results.

\section{Novel Dataset Generation Strategies for Factuality Self-Evaluation}
\label{sec:our_dataset}


This section introduces two novel dataset generation strategies to investigate LLM factuality self-evaluation. To contextualize our approach, we first examine the limitations of the True-False dataset by \citet{azaria2023llmlying}. While it provides a structured framework for evaluating LLMs, its construction imposes constraints that may limit generalizability of the findings. Specifically, true and false statements are derived from tabular data using predefined templates. We argue that this approach suffers from several limitations:
\begin{itemize}[itemsep=1pt, parsep=1pt, topsep=1pt, partopsep=1pt, align=left, leftmargin=*]
    \item \textbf{Adherence to predefined templates:} The use of fixed templates limits the linguistic expressiveness of the dataset, potentially constraining the probe classifier’s ability to generalize beyond rigidly structured statements (e.g., \textit{<company> operates in the industry of <industry>}). 
    \item \textbf{Distribution misalignment:} The statements are constructed from tabular data rather than generated by the LLM itself. As a result, the dataset may not align with the LLM's generative distribution. For instance, a niche true fact in the dataset --but unknown by the model-- could have high perplexity for the LLM, undermining the study's core premise: evaluating an LLM's intrinsic ability to "judge" its false claims.
   
    \item \textbf{Lack of consideration for the LLM’s knowledge state:} LLMs exhibit uneven factual knowledge based on their pretraining data, with strengths in some domains and gaps in others. A model can assess a statement factuality only if it has prior exposure to it. The dataset does not account for these inconsistencies: it evaluates whether an LLM can detect factual errors without considering whether the model actually possesses knowledge of the fact.
  
    \item \textbf{Differences in cardinality:} Some properties in the dataset have fewer admissible values, making certain facts easier to evaluate. For example, a statement like \textit{"<element> appears in its standard state as \_"} has fewer possible values (i.e.,\textit{\{Solid, Liquid, Gas\}}) compared to statements like \textit{"<city> is a city in \_"} that involve a broader set. This imbalance in complexity may bias the evaluation process.
\end{itemize}  
These limitations impact the interpretability of results when evaluating an LLM's internal representations of factuality.
To ensure a more realistic assessment of an LLM's self-evaluation of factuality, we propose two strategies to address these issues. The first strategy samples statements from tabular data to better align with the LLM's generative predictions. The second strategy involves sampling LLM-generated facts as answers to questions from a well-known Question Answering dataset.


\subsection{Perplexity-based Dataset Construction} 
\label{sec:our_dataset_perplexity} 
This section presents a novel dataset generation strategy to address limitations in the \citet{azaria2023llmlying} True-False dataset, particularly distribution misalignment and implausible negative samples. To improve negative sampling based on random property-object substitutions, we introduce a \textbf{perplexity-guided probabilistic sampling} method, which re-weights false statements based on perplexity for better alignment with LLM output distributions. Since perplexity depends on the model, the same LLM under evaluation must generate false statements to ensure consistency, resulting in a model-dependent dataset tailored to the characteristics and biases of the LLM being studied.

True statements are initially constructed, as in \citet{azaria2023llmlying}, by directly inserting correct entity-property pairs into pre-defined sentence templates. Regarding the false statements generation, we proceed as follows: (i) for each true statement, all unique alternative property values are gathered from the entire dataset; (ii) a candidate sentence is created for each alternative property by inserting it into the corresponding template; (iii) the target LLM (i.e., \opt{} or \llamatwo) computes the perplexity of each candidate sentence. This perplexity score serves as a plausibility metric, with lower perplexity indicating a more plausible (yet incorrect) statement.

Given a true statement, we define \( C \) as the set of potential candidate sentences, which includes the true statement. Furthermore, \( C' \subset C \) is defined as the subset of candidate false sentences (i.e. excluding the true statement). Candidates \( c \in C \) are ranked based on their perplexity scores, with lower scores indicating higher plausibility.

Since perplexity can be interpreted as a measure of plausibility, we operate under the assumption that an LLM possesses factual knowledge about a fact if the fact is assigned a "sufficiently low perplexity". Conversely, a high perplexity score for the true statement suggests that the LLM lacks knowledge of the fact.
Given the limitations discussed earlier, we aim to evaluate the LLM ability to discern between true and false statements when it possesses the relevant knowledge. Therefore, we exclude instances where the LLM exhibits limited knowledge about the true statement: 
if the true statement does not rank among the lowest \( k \) perplexity candidates, the generation of that instance is discontinued, and the next true fact is considered. In practice, we define $k$ as $k = \alpha|C|$, where $\alpha$ is a hyperparameter ($0 < \alpha < 1$). This accounts for the varying cardinality of the property ranges in the dataset, ensuring that the threshold for "sufficiently low perplexity" is adjusted based on the number of possible values for a given property.

In addition, we want to simulate a real hallucination scenario where the LLM is uncertain between the true fact and plausible alternatives: given the perplexity score function $PP(\cdot)$, all false candidates $c$ with a perplexity score \(PP(c) < (1+\beta) PP(true)\), where $0<\beta <1$ is a hyperparameter, are considered, resulting in a reduced set of candidates $C^\star$. A min-max normalization is applied to their perplexity,
\begin{equation}
    \text{NormPP}(c) = \frac{\text{PP}(c) - \min_{c \in C} \text{PP}(c)}{\max_{c \in C} \text{PP}(c) - \min_{c \in C} \text{PP}(c)}, 
\end{equation}

The normalized perplexities are transformed using a plausibility score function $s(\cdot)$, i.e., lower perplexity scores result in higher plausibility scores. The scores are then normalized to ensure that they sum to 1 and are treated as a probability distribution over the candidates:
\begin{equation}
    s(c) = e^{-\text{NormPP}(c)}, \quad P(c_i) = \frac{s(c_i)}{\sum_{c_j \in C^\star} s(c_j)}.
\end{equation}
The normalization guarantees that \( P(c_i) \) values are suitable for sampling. Finally, a mixture of top-k and nucleus sampling~\cite{DBLP:conf/iclr/HoltzmanBDFC20} is employed to sample the candidate for insertion into the template. Specifically, we apply top-k and nucleus sampling sequentially: we select the top-k highest-scoring candidates and then refine this set by choosing the smallest subset whose cumulative probability reaches a predefined threshold.
This process generates a coherent yet factually incorrect statement that is more realistic and closely aligned with the LLM internal token prediction patterns. In Section~\ref{sec:dataset-generation-setup} we detail the selected values for the hyperparameters $\alpha$, $\beta$, $k$, and $p$.


\subsection{LLM-Generated Dataset Construction}
\label{sec:our_dataset_trivia}


The strategy in Section~\ref{sec:our_dataset_perplexity} constructs a balanced true-false dataset from tabular data but has inherent limitations. While enabling direct comparison with \citet{azaria2023llmlying}, it restricts generative models to rigid templates, limiting expressiveness. Additionally, its reliance on fixed candidate sets can lead to easily classifiable false statements when the set is small. Finally, template-based sampling reduces diversity, likely due to token bias: we found LMs consistently assigning lower perplexity to certain property-object pairs, regardless of the subject.
To more realistically assess an LLM factuality self-evaluation, we propose generating both true and false statements directly from the model, overcoming the limitations of template-based approaches. This involves (a) a method to elicit diverse factoid statements from the LLM, and (b) a strategy to annotate statement veracity, addressing biases and inconsistencies in the process.
Consider a Question Answering dataset composed of $N$ questions, $D_{QA} = \{(q_i, a_i) \}_{i=1}^{N}$, where each question $q_i$ has a corresponding ground-truth answer $a_i$. Given a LLM $\mathcal{M}$, we prompt it with each question $K$ times, yielding a set of generated answers. This results in an extended dataset $D^\mathcal{M}_{QA} = \{(q_i, a_i, \{a^\mathcal{M}_{i,k}\}^K_{k=1}) \}_{i=1}^{N}$.
Following LLM-as-judge~\cite{DBLP:journals/corr/abs-2411-15594, DBLP:journals/corr/abs-2501-10970} practices, the LLM-generated answers in $D^\mathcal{M}_{QA}$ can be annotated using an oracle LLM, which we assume is able to evaluate the veracity of each answer $a^\mathcal{M}_{i,k}$ given the ground-truth answer $a_i$ and the question $q_i$. This operation results in an annotated dataset
\begin{equation}
\hat{D}^\mathcal{M}_{QA} = \{(q_i, a_i, \{a^\mathcal{M}_{i,k}, v^\mathcal{M}_{i,k}\}_{k=1}^{K}) \}_{i=1}^{N},
\end{equation}
where $v_{i,k}$ is a veracity label assigned by the oracle, indicating whether the generated answer $\hat{a}_{i,k}$ is correct ($v_{i,k} = 1$) or incorrect ($v_{i,k} = 0$).




Increasing K enhances the reliability of responses by offering a more accurate evaluation of the LLM knowledge state regarding a question. This evaluation can lead to three possible outcomes: (i) a high proportion of correct answers suggests the LLM fully understands the facts; (ii) a high proportion of incorrect answers indicates the LLM lacks or has partial knowledge, preventing correct responses; (iii) a mix of correct and incorrect answers implies knowledge with a tendency toward hallucination. This study assumes that an LLM can only encode factuality regarding a generated fact if it has some knowledge about it. Therefore, we focus exclusively on the third scenario. This scenario also naturally produces a balanced dataset, including true and false variations of the same fact.
We define the correct answer ratio $p^\mathcal{M}_i = \frac{1}{K} \sum_{k=1}^{K} v^\mathcal{M}_{i,k}$. We consider questions whose $p_i$ is around 0.5 with a tolerance hyperparameter $\tau$, that is $| p^\mathcal{M}_i - 0.5 | < \tau$.
The dataset is obtained by selecting the answers and their veracity labels satisfying the condition:
\begin{equation}
    D^\mathcal{M}_{Facts} = \{ (a^\mathcal{M}_{i,k}, v^\mathcal{M}_{i,k}) : | p^\mathcal{M}_i - 0.5 | < \tau \}_{i=1}^{N}.
\end{equation}



\section{Experiments and Discussion} 
\label{sec:experiments}

This section describes the experimental setup used to extend the prior investigation,
presents the results, and discusses their implications.
We address the following Research Questions:
\begin{itemize}[itemsep=1pt, parsep=1pt, topsep=1pt, partopsep=1pt, align=left, leftmargin=*]
    \item[\textbf{RQ1}] \textit{Are probing classifiers trained as in \citet{azaria2023llmlying} capable of generalizing to True-False sentences with similar perplexity?}
    \item[\textbf{RQ2}] \textit{Can the same probes generalize to facts generated by LLMs?}
\end{itemize}

\subsection{Dataset Generation Setup}
\label{sec:dataset-generation-setup}
To generate the datasets for training the probe and alleviate the limitations of \citet{azaria2023llmlying}, 
we employ the perplexity-based sampling procedure described in Section~\ref{sec:our_dataset_perplexity}. 
Specifically, we set the needed hyperparameters as follows: $\alpha = 0.1$, $\beta = 0.1$, $k=10$ and $p=0.9$.
We excluded the \textit{Scientific Facts} topic from our perplexity-based sampling procedure, as its original version was generated by ChatGPT and not from tabular data (additional details in Appendix \ref{app:perplexity-dataset}). Table \ref{tab:our_dataset} summarizes the statistics for the two refined True-False datasets, while Figure \ref{fig:radar_comparison} shows the differences in average perplexity between the original and the proposed datasets.

\begin{table}[t]
\centering
\resizebox{\columnwidth}{!}{
    \begin{tabular}{lcccc}
    \toprule
    \multirow{2}{*}{\textbf{Dataset}} &
    \multicolumn{2}{c}{\textbf{\llamatwo{}}} &
    \multicolumn{2}{c}{\textbf{\opt{}}} \\
    \cmidrule{2-5}
    & \textbf{Sentences} & \textbf{(\%) True} & \textbf{Sentences} &  \textbf{(\%) True} \\
    \midrule
    \textbf{Cities} & 674 & 50 & 756 & 50 \\
    \textbf{Inventions} & 336 & 50 & 202 & 50 \\
    \textbf{Elements} & 118 & 50 & 220 & 50 \\
    \textbf{Animals} & 116 & 50 & 114 & 50 \\
    \textbf{Companies} & 326 & 50 & 310 & 50 \\ 
    \bottomrule
    \end{tabular}}
    \caption{Novel generation of our True-False dataset, following the approach described in Section \ref{sec:our_dataset_perplexity}, including number of sentences and percentage of true samples.}
    \label{tab:our_dataset}
\end{table}
\newenvironment{customlegend}[1][]{%
    \begingroup
    \csname pgfplots@init@cleared@structures\endcsname
    \pgfplotsset{#1}%
}{%
    \csname pgfplots@createlegend\endcsname
    \endgroup
}%
\def\addlegendimage{\csname pgfplots@addlegendimage\endcsname}

\begin{figure}[h] 

\centering
\begin{minipage}{0.46\linewidth} 
    \centering
    \begin{tikzpicture}
        \begin{polaraxis}[
            width=\linewidth,
            grid=major,
            xtick={0,72,144,216,288}, 
            xticklabels={Cities, Inventions, Elements, Animals, Companies}, 
            ytick={20, 40, 60, 80, 100}, 
            ymin=0, ymax=100,
            enlargelimits=false,
            axis line style={transparent}, title={\textbf{\small{\llamatwo}}},
            tick label style={font=\tiny}, 
            label style={font=\tiny}, 
            ytick style={draw=none},
            yticklabel style={font=\tiny}, 
            yticklabels={,,60,,100}, 
        ]

        \addplot[color=blue, dotted, thick] coordinates {
            (0,24.2) (72,39.8) (144,35.0) (216,44.9) (288,26.0) (360,24.2)
        };

        \addplot[color=blue] coordinates {
            (0,23.4) (72,24.8) (144,18.0) (216,14.7) (288,22.8) (360,23.4)
        };

        \addplot[color=red, dotted, thick] coordinates {
            (0,44.2) (72,80.6) (144,84.9) (216,78.9) (288,54.2) (360,44.2)
        };

        \addplot[color=red] coordinates {
            (0,22.2) (72,21.6) (144,17.2) (216,14.3) (288,22.0) (360,22.2)
        };

        \end{polaraxis}
    \end{tikzpicture}
\end{minipage}
    %
\begin{minipage}{0.46\linewidth} 
    \centering
    \begin{tikzpicture}
        \begin{polaraxis}[
            width=\linewidth,
            grid=major,
            xtick={0,72,144,216,288}, 
            xticklabels={Cities, Inventions, Elements, Animals, Companies}, 
            ytick={30, 60, 90, 120, 150, 180}, 
            ymin=0, ymax=180,
            enlargelimits=false,
            title={\small{\textbf{\opt}}},
            tick label style={font=\tiny}, 
            label style={font=\tiny}, 
            ytick style={draw=none},
            yticklabel style={font=\tiny}, 
            yticklabels={, , 90, , , 180},
            axis line style={transparent},  
        ]

        \addplot[color=blue, dotted, thick] coordinates {
            (0,31.3) (72,113.8) (144,103.9) (216,115.8) (288,55.3) (360,31.3)
        };

        \addplot[color=blue] coordinates {
            (0,39.1) (72,44.1) (144,95.4) (216,81.2) (288,45.5) (360,39.1)
        };

        \addplot[color=red, dotted, thick] coordinates {
            (0,61.4) (72,160.2) (144,141.4) (216,168.0) (288,103.5) (360,61.4)
        };

        \addplot[color=red] coordinates {
            (0,36.1) (72,40.7) (144,96.7) (216,69.8) (288,41.5) (360,36.1)
        };

        \end{polaraxis}
    \end{tikzpicture}
\end{minipage}
\begin{minipage}{\linewidth} 
    \centering
\begin{tikzpicture}
    \begin{customlegend}[
        legend entries={\tiny{Original PP(T)}, \tiny{Ours PP(T)}, \tiny{Original PP(F)}, \tiny{Ours PP(F)}},
        legend style={column sep=0.8ex},
        legend columns=-1,
        legend image post style={scale=0.6}
    ]
        \addlegendimage{color=blue, dotted, thick}
        \addlegendimage{color=blue}
        \addlegendimage{color=red, dotted, thick}
        \addlegendimage{color=red}
    \end{customlegend}
\end{tikzpicture}
\end{minipage}

    \caption{Comparison of average perplexity scores for \llamatwo{} and \opt{} for the Original dataset by \citet{azaria2023llmlying} and our refined version. Lower perplexity indicates that the sentences are more likely to have been generated by the model. PP(T/F) denotes average perplexity of true/false sentences.}
    \label{fig:radar_comparison}
\end{figure}
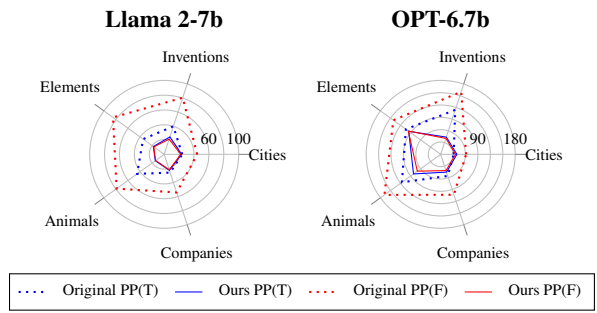

We construct an LLM-generated dataset following the procedure described in Section \ref{sec:our_dataset_trivia}. Our experiments use TriviaQA~\cite{triviaqa}, a dataset of question-answer pairs collected from 14 trivia and quiz-league websites. Given the limited computational resources, we limit our focus to the validation split of TriviaQA Wikipedia questions. We construct a True-False dataset from two additional QA datasets, SQuAD 2.0~\cite{squad_citation} and TruthfulQA ~\cite{truthful_citation}, employing the same approach, which we describe in Appendix~\ref{app:llm_generated_different}.
For each LLM in our analysis, we set $K=10$ and $\tau=0.1$, and we discard the answers composed of less than 5 tokens to improve self-consistency.
For annotation, we use \textbf{GPT-4o mini} with a 3-shot learning strategy, where three examples are provided.
More details are included in Appendix~\ref{app:answer-generation}, and Table~\ref{tab:trivia_dataset} reports the dataset statistics. 
\begin{table*}[t]
    \centering
    \resizebox{\textwidth}{!}{%
    \begin{tabular}{cccccccccccccc}
        \toprule
        \multirow{2}{*}{\textbf{Layer}} & \multirow{2}{*}{\textbf{\makecell[c]{Training\\Data}}}
        & \multicolumn{2}{c}{\textbf{Cities}} 
        & \multicolumn{2}{c}{\textbf{Inventions}} 
        & \multicolumn{2}{c}{\textbf{Elements}} 
        & \multicolumn{2}{c}{\textbf{Animals}} 
        & \multicolumn{2}{c}{\textbf{Companies}} 
        & \multicolumn{2}{c}{\textbf{Average}} \\ 
        \cmidrule(l){3-14}
        & & \llamatwo{} & \opt{} 
        & \llamatwo{} & \opt{} 
        & \llamatwo{} & \opt{} 
        & \llamatwo{} & \opt{} 
        & \llamatwo{} & \opt{} 
        & \llamatwo{} & \opt{} \\ 
        \midrule
        \multirow{2}{*}{\textbf{last}} & Orig. & 0.6882 & 0.5724 & 0.6409 & 0.5094 & 0.6314 & 0.5482 & 0.5685 & 0.5259 & 0.6290 & 0.6984 & 0.6316 & 0.5709 \\ 
                                       & Novel & 0.6365 & 0.6143 & 0.6101 & 0.5144 & 0.5623 & 0.5293 & 0.5461 & 0.4846 & 0.7414 & 0.7211 & 0.6311 & 0.5720 \\ 
        \midrule
        \multirow{2}{*}{\textbf{28}} & Orig. & 0.7056 & 0.5870 & 0.7001 & 0.5178 & 0.6161 & 0.5832 & 0.6013 & \textbf{0.5566} & 0.7079 & 0.7234 & 0.6662 & 0.5936 \\ 
                                      & Novel & 0.5091 & 0.6057 & 0.6591 & 0.5473 & 0.5665 & 0.5655 & 0.6052 & 0.4662 & 0.7061 & 0.7065 & 0.6092 & 0.5811 \\ 
        \midrule
        \multirow{2}{*}{\textbf{24}} & Orig. & 0.8286 & 0.7026 & 0.7250 & 0.6022 & 0.6432 & 0.5918 & 0.6310 & 0.5439 & 0.7121 & 0.7366 & 0.7080 & 0.6354 \\ 
                                      & Novel & 0.6025 & 0.6710 & 0.6609 & 0.6248 & 0.5763 & 0.5993 & 0.5836 & 0.4868 & 0.7868 & 0.7366 & 0.6420 & 0.6225 \\ 
        \midrule
        \multirow{2}{*}{\textbf{20}} & Orig. & 0.8272 & 0.7313 & 0.7741 & \textbf{0.6230} & 0.6492 & \textbf{0.6255} & 0.5832 & 0.5075 & 0.7583 & 0.7502 & 0.7184 & \textbf{0.6475} \\ 
                                      & Novel & 0.7382 & \textbf{0.7528} & 0.6973 & 0.6131 & 0.6051 & 0.5986 & 0.6190 & 0.4807 & \textbf{0.8270} & \textbf{0.7566} & 0.6973 & 0.6404 \\ 
        \midrule
        \multirow{2}{*}{\textbf{16}} & Orig. & 0.8941 & 0.6433 & 0.7888 & 0.5698 & \textbf{0.6801} & 0.5930 & 0.5836 & 0.3816 & 0.7768 & 0.7426 & 0.7447 & 0.5860 \\ 
                                      & Novel & \textbf{0.9301} & 0.7505 & \textbf{0.7961} & 0.5644 & 0.6623 & 0.5568 & \textbf{0.6319} & 0.5307 & 0.8265 & 0.7231 & \textbf{0.7694} & 0.6151 \\ 
        \bottomrule
    \end{tabular}%
    }
    \caption{Accuracy values obtained training SAPLMA on the original True-False dataset and on our refined version, then \textbf{tested on our refined version}. Results are shown for the \llamatwo{} and \opt{} models. `Orig.' denotes the `original True-False dataset as training data, while `Novel' denotes our version of the True-False dataset as training data. In \textbf{bold} we denote the best combination of layer/training dataset for each combination of model/topic.}
    \label{tab:truefalsenew}
\end{table*}

\begin{table}[t]
\resizebox{\columnwidth}{!}{%
\setlength{\tabcolsep}{2pt}
\begin{tabular}{llccccc|ccccc}
\hline
\multirow{2}{*}{\textbf{Dataset}} &
   &
  \multicolumn{5}{c}{\textbf{Threshold = 0.5}} &
  \multicolumn{5}{c}{\textbf{Optimal Threshold}} \\ \cline{3-12} 
 &
   &
  \textbf{last} &
  \textbf{28} &
  \textbf{24} &
  \textbf{20} &
  \textbf{16} &
  \textbf{last} &
  \textbf{28} &
  \textbf{24} &
  \textbf{20} &
  \textbf{16} \\ \hline
\multirow{2}{*}{\textbf{billturnbull}}  & Llama & .561 & .576 & .618 & .621 & .628 & .547 & .581 & .618 & .634 & .639 \\
                                        & OPT   & .547 & .558 & .591 & .551 & .537 & .530 & .558 & .547 & .530 & .499 \\ \hline
\multirow{2}{*}{\textbf{derby*}}        & Llama & .568 & .581 & .575 & .596 & .617 & .560 & .565 & .559 & .571 & .597 \\
                                        & OPT   & .553 & .564 & .584 & .587 & .572 & .562 & .566 & .583 & .586 & .580 \\ \hline
\multirow{2}{*}{\textbf{quiz4free}}     & Llama & .564 & .547 & .523 & .561 & .589 & .548 & .523 & .521 & .547 & .573 \\
                                        & OPT   & .559 & .559 & .575 & .581 & .560 & .544 & .530 & .533 & .546 & .541 \\ \hline
\multirow{2}{*}{\textbf{quizguy}}       & Llama & .578 & .585 & .588 & .607 & .635 & .576 & .587 & .601 & .595 & .637 \\
                                        & OPT   & .579 & .583 & .589 & .590 & .584 & .559 & .555 & .565 & .559 & .548 \\ \hline
\multirow{2}{*}{\textbf{triviabug}}     & Llama & .494 & .518 & .521 & .525 & .538 & .508 & .500 & .530 & .542 & .545 \\
                                        & OPT   & .620 & .624 & .607 & .596 & .528 & .632 & .635 & .605 & .598 & .553 \\ \hline
\multirow{2}{*}{\textbf{businessballs}} & Llama & .566 & .558 & .565 & .574 & .582 & .558 & .551 & .555 & .564 & .575 \\
                                        & OPT   & .559 & .558 & .578 & .570 & .553 & .545 & .547 & .573 & .562 & .551 \\ \hline
\multirow{2}{*}{\textbf{jetpunk}}       & Llama & .587 & .627 & .620 & .643 & .654 & .543 & .580 & .550 & .554 & .601 \\
                                        & OPT   & .606 & .612 & .614 & .596 & .621 & .618 & .618 & .621 & .605 & .619 \\ \hline
\multirow{2}{*}{\textbf{odquiz}}        & Llama & .551 & .536 & .546 & .562 & .573 & .537 & .525 & .532 & .550 & .560 \\
                                        & OPT   & .560 & .573 & .583 & .583 & .542 & .551 & .566 & .571 & .569 & .521 \\ \hline
\multirow{2}{*}{\textbf{quiz-zone}}     & Llama & .556 & .557 & .558 & .565 & .611 & .541 & .549 & .555 & .569 & .615 \\
                                        & OPT   & .569 & .570 & .582 & .592 & .552 & .537 & .550 & .535 & .580 & .508 \\ \hline
\multirow{2}{*}{\textbf{quizballs}}     & Llama & .603 & .575 & .572 & .578 & .571 & .602 & .561 & .561 & .570 & .553 \\
                                        & OPT   & .558 & .565 & .574 & .571 & .540 & .550 & .549 & .564 & .564 & .539 \\ \hline
\multirow{2}{*}{\textbf{quizwise}}      & Llama & .560 & .565 & .579 & .609 & .618 & .563 & .551 & .574 & .605 & .619 \\
                                        & OPT   & .560 & .563 & .565 & .577 & .540 & .555 & .556 & .552 & .575 & .537 \\ \hline
\multirow{2}{*}{\textbf{sfquiz}}        & Llama & .568 & .554 & .554 & .559 & .575 & .564 & .554 & .559 & .566 & .583 \\
                                        & OPT   & .584 & .591 & .589 & .590 & .547 & .570 & .582 & .577 & .573 & .536 \\ \hline
\multirow{2}{*}{\textbf{triviacountry}} & Llama & .536 & .554 & .559 & .556 & .566 & .506 & .530 & .536 & .524 & .549 \\
                                        & OPT   & .536 & .551 & .550 & .587 & .534 & .449 & .476 & .472 & .511 & .472 \\ \hline
\multirow{2}{*}{\textbf{wrexham**}}     & Llama & .570 & .563 & .569 & .553 & .565 & .567 & .548 & .561 & .545 & .566 \\
                                        & OPT   & .548 & .573 & .578 & .584 & .554 & .535 & .552 & .574 & .579 & .571 \\ \hline
\multirow{2}{*}{\textbf{Average}}       & Llama & .562 & .564 & .568 & .579 & .594 & .552 & .550 & .558 & .567 & .587 \\
                                        & OPT   & .567 & .575 & .583 & .583 & .555 & .553 & .560 & .562 & .567 & .587 \\ \hline
\multicolumn{11}{l}{\textit{*: derby is adopted as abbreviation of derbyshirepubquizleague}}                            &       \\
\multicolumn{11}{l}{\textit{**: wrexham is adopted as abbreviation of wrexhamquizleague}}                               &      
\end{tabular}
}
\caption{Performance of SAPLMA on a fact dataset generated from TriviaQA. The original topic-wise leave-one-out strategy is adopted. Results are shown for the \llamatwo{} and \opt{} models.}
\label{tab:trivia}
\vspace{-0.4cm}
\end{table}

\subsection{Impact of Perplexity-Based Sampling on SAPLMA Accuracy}
We assess the impact of our perplexity-based sampling strategy by training the SAPLMA probe classifier separately on the original True-False dataset from \citet{azaria2023llmlying} and our refined version. As shown in Table~\ref{tab:truefalsenew}, classifiers tested on our refined dataset achieve the highest accuracy in deeper layers. For \llamatwo{}, Layer 16 shows the most consistent results, achieving the highest average accuracy across all tested configurations when trained on our dataset. 
Although accuracy fluctuates across different topics, overall performance remains largely consistent between the training configurations. Comparing the results with the original results in Table \ref{tab:sampla_reprod-results}, it emerges that classifiers trained on the original dataset achieve higher accuracy when tested on the same dataset.
This is particularly evident in the ‘Animals’ topic.
However, it is important to consider that the perplexity values in our refined dataset are significantly lower than those in the original dataset for both true and false statements (see~\ref{fig:radar_comparison}). Additionally, our refined dataset features lower and more closely aligned perplexity scores between true and false statements, with false statements often yielding even lower perplexity values than true ones.
This finding implies the validity of \citeauthor{azaria2023llmlying} hypothesis, which states that \textbf{factuality information is encoded in LLM hidden states}, although it might not be immediately evident from the model predicted probabilities. 

\vspace{0.2em}
\textbf{RQ1}: \textit{Are probing classifiers trained as in \citet{azaria2023llmlying} capable of generalizing to True-False sentences with similar perplexity?} 

Surprisingly, the results between the probes trained on our refined dataset and the probes trained on the original dataset are mostly comparable. This proves that \textbf{the probes can generalize even when the train-test datasets have different perplexity}.

\subsection{Generalization of SAPLMA on LLM-generated Sentences}\label{sec:llm-generated_results}
We extend our experiments with the SAPLMA classifier to a new setting, where both training and evaluation are conducted on sentences generated by an LLM and sourced from TriviaQA, leveraging the procedure described in Section~\ref{sec:our_dataset_trivia}.
Table~\ref{tab:trivia} reports the performance of SAPLMA on a set of factual statements extracted from TriviaQA. The results indicate that the currently used probes are inadequate for factuality self-assessment in real-world scenarios, as the observed accuracy does not reach a noteworthy threshold. Furthermore, following the suggestion of \citeauthor{azaria2023llmlying}, we optimize the classification threshold; however, this adjustment yields no significant improvement in performance. For completeness, the interested reader may find the result obtained on the same dataset when training on the original True-False dataset of \citeauthor{azaria2023llmlying} in Appendix \ref{app:original_trivia}. In addition, we report results on SQuAD 2.0 and TruthfulQA in Appendix~\ref{app:llm_generated_different}, which align closely with the results observed on TriviaQA, further supporting the consistency of our findings.

\vspace{0.2em}
\textbf{RQ2}: \textit{Can the same probes generalize to facts generated by LLMs?} 

In summary, this experiment partially contradicts the findings of \citet{azaria2023llmlying}: \textbf{the trained probes are not capable of providing good generalization to an LLM-generated dataset}, even when the accuracy threshold is tuned. The motivation could stem from the nature of the dataset, as TriviaQA contains open-domain questions that result in more nuanced facts than the ones in the original True-False dataset (Section~\ref{sec:replication}) or the OPT-generated dataset (Section~\ref{sec:3-generated}).

\textit{We believe that further research is needed to enhance the effectiveness of factuality self-assessment techniques, particularly in settings involving LLM-generated content. 
Promising research directions may be leveraging datasets that are closely aligned with the distribution of LLM-generated text and exploring alternative techniques such as uncertainty-aware classification.}

\section{Related Work}\label{sec:related} 



Probing techniques have become central for the layer-wise interpretation of deep learning models \cite{alain2016understanding}.
This approach was then extended to Large Language Models (LLMs) to assess LLMs ability to encode syntax and semantics~\cite{conneau2018you,tenney2019you}. 
Specifically, among the different applications, a technique that emerges is \textbf{self-evaluation}, defined as \textit{a model's ability to assess the accuracy of its own outputs}~\cite{kadavath2022}. Among the various works, notable research by \citet{kadavath2022} explored estimating a well-calibrated \textit{P(True)} directly from output probabilities to reflect answer accuracy. \citet{azaria2023llmlying} demonstrate that LLMs are capable of detecting false claims in synthetic true-false datasets. Building upon this initial study, several works~\cite{DBLP:journals/corr/abs-2310-06824, DBLP:conf/nips/BurgerHN24, levinstein2024still} investigate the generalization capabilities of probes in detecting hallucinations in LLMs. While these studies offer valuable insights into how factual knowledge is encoded in LLMs, they stop short of examining whether LLMs can assess the factuality of their own generations. Instead, they primarily evaluate models on artificially constructed or perturbed datasets, rather than on claims produced by the LLMs themselves. \citet{orgad2024} explore error detection and hallucination mitigation in Question Answering by probing the internal representations of question-answer pairs. Their study shows that while LLMs encode factuality signals, these signals do not generalize well across task-specific datasets, suggesting that factuality encoding is task-dependent rather than universal. \citet{gekham} show that LLMs can internally encode knowledge, detectable by probing the hidden states, yet still fail to express it in their generated output. 
\citet{chen2024inside} introduce the INSIDE framework, which detects hallucinations using the EigenScore metric to measure self-consistency across multiple LLM outputs for a single input.
Similarly,~\citet{zhang2024transferable} leverage a probe model, PINOSE, trained via offline consistency-checking, to perform online hallucination detection. Although INSIDE and PINOSE primarily focus on self-consistency in internal representations across multiple generated responses, they do not interpret hidden states or look for explicit factuality encoding. 

\section{Conclusion and Future Work}
\label{sec:conclusion}

In this paper, we investigate the factuality-encoding capabilities of LLMs. Our work replicates the methodology of \citet{azaria2023llmlying} to ensure reproducibility and extend their approach with two novel dataset construction strategies: perplexity-based negative sampling and fact generation based on QA datasets. We applied these strategies to analyze two open-source LLMs, and found that although the findings from previous studies are partially validated even on more challenging synthetic datasets, transferring these findings to LLM-generated datasets proves difficult. This study paves the way for more reliable LLM evaluations and offers practical guidelines for improving model transparency and trustworthiness in real-world applications.

\section*{Limitations}

There are a few aspects of our study that could be explored further. We made every effort to check the datasets for inconsistencies, but a more thorough manual verification by human annotators would be beneficial for ensuring their robustness and minimizing potential biases. Additionally, our analysis is based on a limited set of models. While these models provide valuable insights, it is possible that larger or more complex models could demonstrate enhanced performance, particularly in the context of self-evaluation. Future work could expand on this, incorporating a wider range of models to investigate whether scalability can improve results.

Lastly, we highlight important considerations regarding the use of perplexity as a proxy for plausibility in relation to model knowledge. Recent findings suggest that LLMs often assign similar perplexity scores to both seen and unseen sentences~\cite{DBLP:journals/corr/abs-2402-07841}, which may limit the reliability of perplexity in distinguishing between plausible and implausible content. This challenge intersects with broader issues of memorization and data provenance. While our work focuses on extending and refining prior approaches to factuality self-evaluation, we recognize that a rigorous assessment of perplexity effectiveness in this context is a critical avenue for future investigation. We therefore encourage further research into more robust and theoretically grounded measures of plausibility for LLM-generated content.

\section*{Ethical Statement}
A major concern when working with LLMs is their tendency to generate factually inaccurate information. When training probe classifiers to assess factual accuracy, biases and beliefs from the LLM may transfer to the probe, potentially reinforcing cultural, demographic, or ideological biases in factuality self-evaluation. With careful design, probing techniques can be adapted not only to minimize bias but also to actively mitigate its consequences.

\appendix

\newpage
\section*{Appendix}
\section{Dataset Statistics}
\label{sec:appendix-datasets}
\subsection{True-False Dataset~\cite{azaria2023llmlying} Statistics}
Table~\ref{tab:reliable_source} provides a summary of the number of sentences and the distribution of true and false statements across each topic in the original True-False dataset~\cite{azaria2023llmlying} on which we base our replication experiments.


\begin{table}[ht!]
\centering
\resizebox{\columnwidth}{!}{%
    \begin{tabular}{@{}lclc@{}}
    \toprule
    \textbf{Dataset} & \textbf{Sentences} & \textbf{Source} & \textbf{(\%) True} \\ 
    \midrule
    \textbf{Cities} & 1458 & SimpleMaps dataset & 50 \\
    \textbf{Inventions} & 876 & Wikipedia’s list of inventors & 53 \\
    \textbf{Chemical Elements} & 930 & PubChem’s periodic table & 50 \\
    \textbf{Animals} & 1008 & National Geographic Kids & 50 \\
    \textbf{Companies} & 1200 & Forbes Global 2000 List 2022: The Top 200 & 50 \\
    \textbf{Scientific Facts} & 612 & ChatGPT and human annotators & 50 \\ 
    \bottomrule
    \end{tabular}%
}
\caption{True-False dataset categories, original sources, and label splits from~\cite{azaria2023llmlying}.}
\label{tab:reliable_source}
\end{table}

\subsection{LLM-Generated Trivia Facts Dataset}
\label{app:trivia-dataset}
Table~\ref{tab:trivia_dataset} illustrates the number of annotated facts extracted from TriviaQA~\cite{triviaqa} employing the procedure described in section~\ref{sec:our_dataset_trivia}. Statistics are presented for each of the 14 question sources. Table~\ref{tab:dataset_generated_examples} reports examples of generated facts for the \llamatwo{} and \opt{} models.

\begin{table}[ht!]
\centering

\resizebox{\columnwidth}{!}{%
\begin{tabular}{lcccc}
\hline
\multirow{2}{*}{\textbf{Dataset}} & \multicolumn{2}{c}{\textbf{\llamatwo{}}} & \multicolumn{2}{c}{\textbf{\opt{}}} \\
\cline{2-5}
 & \textbf{Sentences} & \textbf{(\%) True} & \textbf{Sentences} & \textbf{(\%) True} \\
\hline
triviacountry & 118 & 49.15\% & 118 & 49.15\% \\
wwwodquiz & 700 & 49.71\% & 525 & 49.71\% \\
triviabug & 97 & 50.52\% & 88 & 51.14\% \\
derby* & 342 & 50.00\% & 325 & 50.00\% \\
quiz-zone & 187 & 48.13\% & 239 & 50.63\% \\
businessballs & 433 & 50.84\% & 443 & 49.43\% \\
wrexham** & 216 & 50.00\% & 276 & 50.72\% \\
sfquiz & 1054 & 49.80\% & 1085 & 50.39\% \\
quizwise & 565 & 49.91\% & 704 & 50.42\% \\
billturnbull & 216 & 49.54\% & 105 & 50.48\% \\
jetpunk & 139 & 51.80\% & 364 & 50.27\% \\
quizballs & 500 & 49.60\% & 488 & 50.82\% \\
quizguy & 240 & 51.67\% & 341 & 50.74\% \\
quiz4free & 171 & 49.12\% & 141 & 48.94\% \\
\hline
\end{tabular}}
\caption{Summary of the dataset obtained by extracting factoid sentences from TriviaQA~\cite{triviaqa} Wikipedia validation split, following the procedure described in Section \ref{sec:our_dataset_trivia}.}
\label{tab:trivia_dataset}
\end{table}

\begin{table}[ht!]
    \centering
   
    \resizebox{\columnwidth}{!}{%
    \setlength{\tabcolsep}{5pt}
    \begin{tabular}{@{}lcl}
    \toprule
    \textbf{Model} & \textbf{Label} & \textbf{Sentence} \\
    \midrule

    \multirow{7}{*}{\llamatwo{}} 
    & 1 & Arthur was married to Guinevere. \\
    & 1 & Arthur’s most famous wife was Guinevere. \\
    & 1 & Guinevere was married to Arthur. \\
    & 1 & Guinevere's husband was King Arthur. \\
    & 0 & Lancelot was married to Guinevere. \\
    & 0 & Sir Lancelot was married to Queen Guinevere. \\
    & 0 & Sir Leonne was married to Queen Guinevere. \\

    \midrule

    \multirow{7}{*}{\opt} 
    & 0 & Anakin Skywalker is Darth Vader's son. \\
    & 0 & Darth Vader's son is Darth Vader. \\
    & 1 & Darth Vader's son is Luke Skywalker, a member of the Rebel Alliance. \\
    & 1 & Darth Vader's son is Luke Skywalker. \\
    & 1 & Darth Vader's son is Luke. \\
    & 0 & Darth Vader's son is known as Darth Vader. \\
    & 1 & Luke's father is Darth Vader. \\

    \bottomrule
    \end{tabular}}
    \caption{Examples of true and false facts generated by \llamatwo{} and \opt{} based on the questions in TriviaQA~\cite{triviaqa}, obtained following the procedure detailed in Section~\ref{sec:our_dataset_trivia}.}
    \label{tab:dataset_generated_examples}
\end{table}

\section{Perplexity-Based Refinement of the True-False Dataset \cite{azaria2023llmlying}}
\label{app:perplexity-dataset}

We base our dataset generation strategy described in Section~\ref{sec:our_dataset} on the same tabular data employed by \citeauthor{azaria2023llmlying}, which was made available upon request.
The properties used for sampling are analogous to the ones employed by the original authors in their dataset generation. However, for the \textit{Cities} topic, we restrict the analysis to facts like "\textit{<city> is city in <country>}." Differently from the original authors, we avoid generating facts such as "\textit{<city> is the name of a city/country}" having only two possible values, potentially resulting in too many easy-to-classify samples.  Table~\ref{tab:perplexity_dataset_examples} reports examples of false sentences generated following the perplexity-based sampling procedure described in Section~\ref{sec:our_dataset_perplexity} and the \llamatwo{} model.

\begin{table}[ht!]
    \centering
    \resizebox{\columnwidth}{!}{%
    \setlength{\tabcolsep}{4pt}
    \begin{tabular}{@{}p{5cm}p{5.5cm}p{5.5cm}@{}}
    \toprule
    \textbf{True Sentence} & \textbf{Original Negative} & \textbf{Generated Negative (\llamatwo{})} \\
    \midrule
    
    The crocodile has a habitat of freshwater. &
    The crocodile has a habitat of \textbf{various.} &
    The crocodile has a habitat of \textbf{grassland/savanna.} \\
    \midrule
    The zebra has distinctive black and white stripes, which may help deter flies and provide camouflage. &
    The zebra is a \textbf{fast swimmer} and can maintain high speeds for extended periods of time. &
    The zebra is the \textbf{fastest land animal}, reaching speeds up to 60-70 mph. \\
    \midrule
    Tantalum has the symbol Ta. &
    Tantalum has the symbol \textbf{Cs.} &
    Tantalum has the symbol \textbf{Tm.} \\
    
    \bottomrule
    \end{tabular}}
    \caption{Examples of false sentences generated with the \llamatwo{} model, using the perplexity-based generation strategy described in Section~\ref{sec:our_dataset_perplexity}. }
    \label{tab:perplexity_dataset_examples}
\end{table}

\section{LLM-Generated Dataset Extraction Details}
\label{app:dataset-extraction}
\subsection{Factoid Answer Generation}
\label{app:answer-generation}
Below is an example prompt used in the text generation pipeline for TriviaQA~\cite{triviaqa}. The 10 examples are sampled from the train Wikipedia split. The answers for the examples are manually crafted by looking at the available ground truth.

\begin{lstlisting}[basicstyle=\ttfamily\footnotesize, breaklines=true]
Question: Where in England was Dame Judi Dench born?
Answer: The English actress Dame Judi Dench was born in York, England.

Question: From which country did Angola achieve independence in 1975?
Answer: Angola achieved independence from Portugal in 1975.

Question: Which city does David Soul come from?
Answer: David Soul hails from Chicago, Illinois.

Question: Who won Super Bowl XX?
Answer: The Chicago Bears won Super Bowl XX.

Question: Which was the first European country to abolish capital punishment?
Answer: Norway was the first European country to abolish capital punishment.

Question: In which country did the widespread use of ISDN begin in 1988?
Answer: The widespread use of ISDN began in Japan in 1988.

Question: What is Bruce Willis' real first name?
Answer: Bruce Willis' real first name is Walter.

Question: Which William wrote the novel Lord of the Flies?
Answer: The William who wrote Lord of the Flies was William Golding.

Question: How is Joan Molinsky better known?
Answer: Joan Molinsky is better known as Joan Rivers.

Question: In which branch of the arts is Patricia Neary famous?
Answer: Patricia Neary is famous in the field of ballet.


\end{lstlisting}

To generate responses, the model is provided with a continuation prompt such as:
\begin{lstlisting}[basicstyle=\ttfamily\footnotesize, breaklines=true]
Question: Ford Prefect came from a star in which constellation?
Answer:
\end{lstlisting}

The model then generates a response, adhering to the pattern set by the previous examples. Responses are filtered based on length and uniqueness to ensure meaningful and high-quality outputs. 

The generation process uses the following hyperparameters for all the employed models (i.e. \llamatwo{} and \opt{}):
\begin{itemize}
\item \texttt{max\_new\_tokens}: 128
\item \texttt{top\_k}: 50
\item \texttt{top\_p}: 0.95
\item \texttt{temperature}: 1.0
\item \texttt{num\_return\_sequences}: 10
\end{itemize}


\subsection{Factoid Answer Annotation}
\label{app:answer-annotation}

Below is an example prompt to \textbf{GPT-4o mini} (2024-08-01), the LLM that we use as the oracle for the procedure described in Section~\ref{sec:our_dataset_trivia}. In this setup, we employ a 3-shot learning strategy, where the model is provided with three examples to establish a clear pattern of judgment. These examples demonstrate the task of evaluating answers against a given ground truth. The model is then tasked with using this context to evaluate a new answer to a question, based on the same criteria. In our experiments, we employ a temperature setting of 1. 

\begin{lstlisting}[basicstyle=\ttfamily\footnotesize, breaklines=true]
SYSTEM:

You are a judge and your role is to judge whether the provided answer is correct for the given question, based on the provided ground truth. Answer with a 1 if the answer is correct and 0 if the answer is incorrect.
Here are a few examples:

USER: 

Question: Who was the next British Prime Minister after Arthur Balfour?
Answer: Arthur Balfour was followed by David Lloyd George.
Ground truth: "{'aliases': ['Sir Henry Campbell-Bannerman', 'Campbell-Bannerman', 'Campbell Bannerman', 'Sir Henry Campbell Bannerman', 'Henry Campbell Bannerman', 'Henry Campbell-Bannerman'], 'normalized_aliases': ['henry campbell bannerman', 'sir henry campbell bannerman', 'campbell bannerman'], 'matched_wiki_entity_name': '', 'normalized_matched_wiki_entity_name': '', 'normalized_value': 'campbell bannerman', 'type': 'WikipediaEntity', 'value': 'Campbell-Bannerman'}"
Evaluation: 

ASSISTANT: 

0

USER:

Question: Who had a 70s No 1 hit with Kiss You All Over?
Answer: The band Exile had a 70s No 1 hit with Kiss You All Over.
Ground truth: "{'aliases': ['Internal exile', 'Exiles', 'Transported for life', 'Exile (politics and government)', 'Voluntary exile', 'Sent into exile', 'Exile and Banishment', 'Self-exile', 'Forced exile', 'Exile', 'Exile in Greek tragedy', 'Banish', 'Banishment'], 'normalized_aliases': ['exiles', 'voluntary exile', 'forced exile', 'banish', 'self exile', 'exile politics and government', 'exile in greek tragedy', 'sent into exile', 'banishment', 'transported for life', 'exile', 'internal exile', 'exile and banishment'], 'matched_wiki_entity_name': '', 'normalized_matched_wiki_entity_name': '', 'normalized_value': 'exile', 'type': 'WikipediaEntity', 'value': 'Exile'}"
Evaluation: 

ASSISTANT: 

1

USER: 

Question: Which common mineral is used to make casts, moulds, blackboard chalk and plaster of Paris?
Answer: The common mineral used to make casts, moulds, blackboard chalk and plaster of Paris is calcium carbonate.
Ground truth: "{'aliases': ['CaSO4.2H2O', 'Gypsum', 'Calcium sulfate dihydrate', 'CaSO4*2H2O', 'Gipsum'], 'normalized_aliases': ['calcium sulfate dihydrate', 'caso4 2h2o', 'gipsum', 'caso4.2h2o', 'gypsum'], 'matched_wiki_entity_name': '', 'normalized_matched_wiki_entity_name': '', 'normalized_value': 'gypsum', 'type': 'WikipediaEntity', 'value': 'Gypsum'}"
Evaluation: 

ASSISTANT:

0
\end{lstlisting}

\section{Additional Experiments on the LLM-Generated Dataset}\label{app:original_trivia}

\begin{table}[ht!]
    \centering
    \resizebox{\columnwidth}{!}{%
    \setlength{\tabcolsep}{2pt}
\begin{tabular}{llccccc|ccccc}

\hline
\multirow{2}{*}{\textbf{Dataset}} &
   &
  \multicolumn{5}{c}{\textbf{Threshold = 0.5}} &
  \multicolumn{5}{c}{\textbf{Optimal Threshold}} \\ \cline{3-12} 
 &
   &
  \textbf{last} &
  \textbf{28} &
  \textbf{24} &
  \textbf{20} &
  \textbf{16} &
  \textbf{last} &
  \textbf{28} &
  \textbf{24} &
  \textbf{20} &
  \textbf{16} \\ \hline
\multirow{2}{*}{\textbf{billturnbull}}            & Llama{} & .579 & .560 & .560 & .593 & .648 & .605 & .553 & .632 & .632 & .691 \\
 &
  OPT &
  .543 &
  .533 &
  .533 &
  .486 &
  .476 &
  .527 &
  .500 &
  .486 &
  .500 &
  .500 \\ \hline
\multirow{2}{*}{\textbf{derby*}} & Llama & .556 & .576 & .550 & .544 & .602 & .542 & .529 & .529 & .554 & .575 \\
 &
  OPT &
  .526 &
  .548 &
  .554 &
  .563 &
  .535 &
  .535 &
  .583 &
  .570 &
  .579 &
  .561 \\ \hline
\multirow{2}{*}{\textbf{quiz4free}} &
  Llama &
  .602 &
  .573 &
  .544 &
  .538 &
  .608 &
  .575 &
  .600 &
  .525 &
  .550 &
  .642 \\
 &
  OPT &
  .525 &
  .489 &
  .511 &
  .525 &
  .504 &
  .545 &
  .535 &
  .556 &
  .576 &
  .495 \\ \hline
\multirow{2}{*}{\textbf{quizguy}} &
  Llama &
  .608 &
  .571 &
  .546 &
  .567 &
  .571 &
  .601 &
  .595 &
  .583 &
  .577 &
  .565 \\
 &
  OPT &
  .557 &
  .557 &
  .587 &
  .569 &
  .557 &
  .586 &
  .594 &
  .552 &
  .548 &
  .552 \\ \hline
\multirow{2}{*}{\textbf{triviabug}} &
  Llama &
  .412 &
  .619 &
  .526 &
  .598 &
  .577 &
  .485 &
  .471 &
  .471 &
  .544 &
  .544 \\
 &
  OPT &
  .557 &
  .602 &
  .671 &
  .614 &
  .500 &
  .629 &
  .597 &
  .629 &
  .597 &
  .532 \\ \hline
\multirow{2}{*}{\textbf{businessballs}} &
  Llama &
  .580 &
  .577 &
  .559 &
  .575 &
  .589 &
  .586 &
  .605 &
  .546 &
  .563 &
  .592 \\
 &
  OPT &
  .564 &
  .578 &
  .555 &
  .587 &
  .521 &
  .585 &
  .579 &
  .585 &
  .547 &
  .537 \\ \hline
\multirow{2}{*}{\textbf{jetpunk}} &
  Llama &
  .612 &
  .590 &
  .583 &
  .619 &
  .640 &
  .582 &
  .561 &
  .520 &
  .592 &
  .673 \\
 &
  OPT &
  .569 &
  .571 &
  .593 &
  .604 &
  .566 &
  .620 &
  .631 &
  .631 &
  .631 &
  .635 \\ \hline
\multirow{2}{*}{\textbf{odquiz}} &
  Llama &
  .546 &
  .536 &
  .537 &
  .559 &
  .564 &
  .527 &
  .522 &
  .547 &
  .563 &
  .571 \\
 &
  OPT &
  .511 &
  .543 &
  .591 &
  .579 &
  .552 &
  .565 &
  .535 &
  .592 &
  .571 &
  .565 \\ \hline
\multirow{2}{*}{\textbf{quiz-zone}} &
  Llama &
  .578 &
  .519 &
  .588 &
  .562 &
  .578 &
  .611 &
  .534 &
  .534 &
  .534 &
  .557 \\
 &
  OPT &
  .544 &
  .577 &
  .603 &
  .611 &
  .586 &
  .542 &
  .518 &
  .554 &
  .601 &
  .613 \\ \hline
\multirow{2}{*}{\textbf{quizballs}} &
  Llama &
  .610 &
  .568 &
  .592 &
  .582 &
  .582 &
  .617 &
  .563 &
  .583 &
  .557 &
  .586 \\
 &
  OPT &
  .535 &
  .578 &
  .559 &
  .594 &
  .549 &
  .532 &
  .564 &
  .512 &
  .599 &
  .576 \\ \hline
\multirow{2}{*}{\textbf{quizwise}} &
  Llama &
  .572 &
  .588 &
  .570 &
  .572 &
  .586 &
  .581 &
  .598 &
  .611 &
  .616 &
  .616 \\
 &
  OPT &
  .540 &
  .550 &
  .574 &
  .568 &
  .551 &
  .580 &
  .550 &
  .554 &
  .582 &
  .576 \\ \hline
\multirow{2}{*}{\textbf{sfquiz}} &
  Llama &
  .530 &
  .533 &
  .528 &
  .538 &
  .560 &
  .545 &
  .581 &
  .584 &
  .570 &
  .575 \\
 &
  OPT &
  .531 &
  .546 &
  .568 &
  .590 &
  .545 &
  .595 &
  .597 &
  .599 &
  .611 &
  .553 \\ \hline
\multirow{2}{*}{\textbf{triviacountry}} &
  Llama &
  .602 &
  .636 &
  .602 &
  .602 &
  .619 &
  .639 &
  .602 &
  .639 &
  .590 &
  .578 \\
 &
  OPT &
  .525 &
  .542 &
  .585 &
  .551 &
  .525 &
  .494 &
  .602 &
  .506 &
  .578 &
  .446 \\ \hline
\multirow{2}{*}{\textbf{wrexham**}} &
  Llama &
  .532 &
  .519 &
  .528 &
  .574 &
  .583 &
  .474 &
  .526 &
  .533 &
  .566 &
  .579 \\
 &
  OPT &
  .500 &
  .533 &
  .594 &
  .583 &
  .525 &
  .572 &
  .552 &
  .562 &
  .608 &
  .567 \\ \hline
\multirow{2}{*}{\textbf{Average}} &
  Llama &
  .566 &
  .569 &
  .558 &
  .573 &
  .593 &
  .569 &
  .560 &
  .560 &
  .572 &
  .596 \\
 &
  OPT &
  .538 &
  .553 &
  .577 &
  .573 &
  .535 &
  .565 &
  .567 &
  .563 &
  .581 &
  .551 \\ \hline
\multicolumn{11}{l}{\textit{*: derby is adopted as abbreviation of derbyshirepubquizleague}}                            &       \\
\multicolumn{11}{l}{\textit{**: wrexham is adopted as abbreviation of wrexhamquizleague}}                               &      
\end{tabular}}%
     
    \caption{Accuracy values obtained training SAPLMA on the original True-False dataset and testing on our facts dataset generated from TriviaQA. The original topic-wise leave-one-out strategy is adopted. Results are shown for the \llamatwo{} and \opt{} models.}
    \label{tab:original_trivia}
\end{table}

Table~\ref{tab:original_trivia} reports the performances of the SAPLMA classifier trained on the original True-False dataset by \citet{azaria2023llmlying} and tested on the dataset generated from TriviaQA. Similarly to Table~\ref{tab:trivia}, SAPLMA does not generalize well over LLM-generated facts. Moreover, tuning an optimal threshold did not provide solid enhancements.

\section{Experiments on Additional LLM-Generated Datasets}
\label{app:llm_generated_different}

We extended our LLM-generated dataset construction approach (Section~\ref{sec:our_dataset_trivia}) to TruthfulQA~\cite{truthful_citation} and SQuAD 2.0~\cite{squad_citation} to further support our findings.

\subsection{LLM-Generated Facts from TruthfulQA}
\begin{table}[h!]
\centering
\resizebox{\columnwidth}{!}{
\begin{tabular}{lcc}
\toprule
\textbf{Model} & \textbf{Total Samples} & \textbf{True Samples (\%)} \\
\midrule
\llamatwo{} & 842 & 49.76\% \\
\opt{}         & 331 & 49.85\% \\
\bottomrule
\end{tabular}}
\caption{Summary of the dataset obtained by extracting factoid sentences from TruthfulQA~\cite{truthful_citation}, following the procedure described in Section~\ref{sec:our_dataset_trivia}.}
\label{tab:truthfulqa_dataset}
\end{table}
\begin{table}[ht]
\resizebox{\columnwidth}{!}{
\begin{tabular}{@{}ccc|cc@{}}
\toprule
 & \multicolumn{2}{c}{\textbf{\opt{}}} & \multicolumn{2}{c}{\textbf{\llamatwo{}}} \\ \midrule
               & Random Split 1    & Random Split 2    & Random Split 1       & Random Split 2      \\ \midrule
\textbf{last}  & 0.487 & 0.488 & 0.491 & 0.504 \\ 
\textbf{28}  & 0.477 & 0.495 & 0.498 & 0.506 \\
\textbf{24}  & 0.478 & 0.513 & 0.507 & 0.507 \\ 
\textbf{20} & 0.512 & 0.535 & 0.518 & 0.509 \\ 
\textbf{16} & 0.493 & 0.492 & 0.520 & 0.514 \\ \bottomrule
\end{tabular}}
\caption{Performance of SAPLMA on a fact dataset generated from TruthfulQA. A 50-50 holdout strategy is adopted. Results are shown for the \llamatwo{} and \opt{} models.}
\label{tab:truthful}
\end{table}
\begin{table}[]
\resizebox{\columnwidth}{!}{
\begin{tabular}{@{}ccc|cc@{}}
\toprule
\textbf{} & \multicolumn{2}{c|}{\textbf{\opt{}}} & \multicolumn{2}{c}{\textbf{\llamatwo{}}} \\
          & Random Split 1     & Random Split 2    & Random Split 1       & Random Split 2      \\ \midrule
\textbf{last} & 0.473 & 0.560 & 0.518 & 0.539 \\
\textbf{28}   & 0.515 & 0.566 & 0.520 & 0.514 \\
\textbf{24}   & 0.503 & 0.476 & 0.518 & 0.501 \\
\textbf{20}   & 0.455 & 0.542 & 0.499 & 0.525 \\
\textbf{16}   & 0.479 & 0.500   & 0.485 & 0.530 \\ \bottomrule
\end{tabular}}
\caption{Accuracy values obtained training SAPLMA on the original True-False dataset and testing on our facts dataset generated from TruthfulQA. A 50-50 holdout strategy is adopted. Results are shown for the \llamatwo{} and \opt{} models.}
\label{tab:truthful_original}
\end{table}

Following the same procedure detailed in Section ~\ref{sec:our_dataset_trivia}, we extract true and false facts from TruthfulQA. Table~\ref{tab:truthfulqa_dataset} reports the statistics of the LLM-generated datasets constructed with \llamatwo{} and \opt{}. Table~\ref{tab:truthful} presents the performance of the SAPLMA classifier when both trained and tested on the TruthfulQA dataset. Additionally, Table~\ref{tab:truthful_original} shows the results obtained by SAPLMA on TruthfulQA after being trained on the original True-False dataset. Since the dataset lacks a topic-based split, we adopt a 50-50 holdout strategy for evaluation. The results on the TruthfulQA dataset support our findings: the probe classifier fails to generalize effectively to LLM-generated datasets.

\subsection{LLM-Generated Facts from SQuAD 2.0}
\begin{table}[]
\centering
\resizebox{\columnwidth}{!}{
\begin{tabular}{llcc}
\toprule
\textbf{Topic} & \textbf{Model} & \textbf{Sentences} & \textbf{(\%) True} \\
\midrule
\multirow{2}{*}{\textbf{Economic\_inequality}} & \llamatwo{} & 224 & 50.45\% \\
                    & \opt{}         & 135 & 50.37\% \\
\midrule
\multirow{2}{*}{\textbf{Immune\_system}}       & \llamatwo{} & 167 & 49.70\% \\
                    & \opt{}         & 72  & 50.00\% \\
\midrule
\multirow{2}{*}{\textbf{Rhine}}               & \llamatwo{} & 120 & 50.83\% \\
                    & \opt{}         & 59  & 47.46\% \\
\midrule
\multirow{2}{*}{\textbf{Warsaw}}              & \llamatwo{} & 140 & 51.43\% \\
                    & \opt{}         & 102 & 50.00\% \\
\midrule
\multirow{2}{*}{\textbf{Yuan Dynasty}}        & \llamatwo{} & 109 & 49.54\% \\
                    & \opt{}         & 53  & 52.83\% \\
\bottomrule
\end{tabular}}
\caption{Summary of the dataset obtained by extracting factoid sentences from SQuAD 2.0~\cite{squad_citation}, following the procedure described in Section~\ref{sec:our_dataset_trivia}.}
\label{tab:squad_datasets}
\end{table}
\begin{table}[h]
\resizebox{\columnwidth}{!}{
\begin{tabular}{@{}ccccccc@{}}
\toprule
                     \textbf{Dataset} & \multicolumn{1}{l}{} & \textbf{last} & \textbf{28} & \textbf{24} & \textbf{20} & \textbf{16} \\ \midrule
\multirow{2}{*}{\textbf{Economic\_inequality}} & Llama                & 0.525         & 0.510        & 0.515       & 0.492       & 0.502       \\
                                               & OPT                  & 0.517         & 0.516       & 0.499       & 0.496       & 0.464       \\ \midrule
\multirow{2}{*}{\textbf{Immune\_system}}       & Llama                & 0.486         & 0.490        & 0.503       & 0.502       & 0.547       \\
                                               & OPT                  & 0.436         & 0.439       & 0.401       & 0.438       & 0.435       \\ \midrule
\multirow{2}{*}{\textbf{Rhine}}                & Llama                & 0.525         & 0.520        & 0.515       & 0.524       & 0.526       \\
                                               & OPT                  & 0.484         & 0.485       & 0.522       & 0.547       & 0.521       \\ \midrule
\multirow{2}{*}{\textbf{Warsaw}}               & Llama                & 0.558         & 0.542       & 0.545       & 0.546       & 0.555       \\
                                               & OPT                  & 0.521         & 0.531       & 0.545       & 0.544       & 0.509       \\ \midrule
\multirow{2}{*}{\textbf{Yuan\_dynasty}}        & Llama                & 0.509         & 0.522       & 0.514       & 0.529       & 0.544       \\
                                               & OPT                  & 0.544         & 0.560        & 0.512       & 0.542       & 0.512       \\ \bottomrule
\end{tabular}}
\caption{Performance of SAPLMA on a fact dataset generated from SQuAD 2.0, restricted to the validation set and the top-5 most popular topics. The original topic-wise leave-one-out strategy is adopted. Results are shown for the \llamatwo{} and \opt{} models.}
\label{tab:squad}
\end{table}
\begin{table}[ht]
\resizebox{\columnwidth}{!}{
\begin{tabular}{@{}ccccccc@{}}
\toprule
                                               & \multicolumn{1}{l}{} & \textbf{last} & \textbf{28} & \textbf{24} & \textbf{20} & \textbf{16} \\ \midrule
\multirow{2}{*}{\textbf{Economic\_inequality}} & Llama                & 0.558         & 0.531       & 0.589       & 0.531       & 0.545       \\
                                               & OPT                  & 0.600           & 0.600         & 0.548       & 0.533       & 0.533       \\ \midrule
\multirow{2}{*}{\textbf{Immune\_system}}       & Llama                & 0.569         & 0.593       & 0.593       & 0.575       & 0.599       \\
                                               & OPT                  & 0.611         & 0.611       & 0.611       & 0.708       & 0.542       \\ \midrule
\multirow{2}{*}{\textbf{Rhine}}                & Llama                & 0.467         & 0.492       & 0.475       & 0.442       & 0.525       \\
                                               & OPT                  & 0.525         & 0.559       & 0.593       & 0.593       & 0.576       \\ \midrule
\multirow{2}{*}{\textbf{Warsaw}}               & Llama                & 0.614         & 0.579       & 0.607       & 0.600         & 0.593       \\
                                               & OPT                  & 0.48          & 0.500         & 0.539       & 0.598       & 0.549       \\ \midrule
\multirow{2}{*}{\textbf{Yuan\_dynasty}}        & Llama                & 0.523         & 0.523       & 0.541       & 0.505       & 0.550        \\
                                               & OPT                  & 0.453         & 0.547       & 0.491       & 0.528       & 0.547       \\ \bottomrule
\end{tabular}}
\caption{Accuracy values obtained training SAPLMA on the original True-False dataset and testing on the facts dataset generated from SQuAD 2.0. Results are shown for the \llamatwo{} and \opt{} models.}
\label{tab:squad_original}
\end{table}
For the SQuAD 2.0 dataset, we limit our dataset construction procedure (Section~\ref{sec:our_dataset_trivia}) to the validation split and the five most common topics, as the large number of questions makes applying our procedure to the full dataset computationally impractical due to time constraints. Table~\ref{tab:squad} presents the performance of the SAPLMA classifier when trained and tested on the SQuAD 2.0 dataset. Table~\ref{tab:squad_original}, instead, reports the results obtained on SQuAD 2.0 after training on the original True-False dataset from \citet{azaria2023llmlying}.

The results on the SQuAD 2.0 dataset further support our conclusion: existing probes fall short in accurately assessing factuality in real-world settings. Across all cases, accuracy remains below a meaningful threshold, highlighting the limited reliability of these methods for self-assessment in scenarios involving LLM-generated sentences.

\section{Examples of SAPLMA Predictions}

The tables in this appendix provide illustrative examples that complement the quantitative results discussed in the main text. Specifically, Table~\ref{tab:examples} presents sentence-level predictions on facts generated from the TriviaQA dataset using Llama 2-7b. In this case, the probe classifier is trained adopting a topic-wise leave-one-out strategy, as in Table~\ref{tab:trivia}. Table~\ref{tab:examples_mitch} shows additional predictions obtained using the probe classifier, which is trained on all but one topic and tested on our refined version of the held-out topic (as in the \textit{Orig.} setting of Table~\ref{tab:truefalsenew}).


\noindent The code is provided in our \texttt{\href{https://github.com/sisinflab/HidingInTheHiddenStates}{GitHub Repository}}.

\begin{table}[t]
\resizebox{\columnwidth}{!}{
\begin{tabular}{@{}p{7cm}ccc@{}}
\toprule
\textbf{Sentence}                                             & \textbf{Predicted}           & \textbf{GT} & \textbf{Confidence} \\ \midrule
Sir John Suckling was the first poet to be buried at Poet’s Corner in London’s Westminster Abbey. & {\color[HTML]{32CB00} False} & False & 0.373 \\
The first poet to be buried at Poet’s Corner in London’s Westminster Abbey was Geoffrey Chaucer.  & {\color[HTML]{32CB00} True}  & True  & 0.551 \\
The poet who was the first to be buried in Westminster Abbey was Geoffrey Chaucer.                & {\color[HTML]{FE0000} False} & True  & 0.456 \\ \midrule
Roquefort cheese is made from sheep's milk.                   & {\color[HTML]{32CB00} True}  & True        & 0.794               \\
Roquefort cheese is made from cow's milk.                     & {\color[HTML]{FE0000} True}  & False       & 0.851               \\
The milk of sheep is used to make 'Roquefort Cheese'.         & {\color[HTML]{FE0000} False} & True        & 0.488               \\ \midrule
South Korea held its first 'Grand Prix' motor race in 1999.   & {\color[HTML]{32CB00} False} & False       & 0.291               \\
South Korea held its first 'Grand Prix' motor race in 2010.   & {\color[HTML]{FE0000} False} & True        & 0.305               \\
South Korea hosted its first 'Grand Prix' motor race in 1966. & {\color[HTML]{32CB00} False} & False       & 0.182               \\ \bottomrule
\end{tabular}}
\caption{Examples of sentence classification on facts generated from the TriviaQA dataset using \llamatwo{}. The original topic-wise leave-one-out strategy is adopted. Correct predictions are highlighted in {\color[HTML]{32CB00} green}, while incorrect ones are shown in {\color[HTML]{FE0000} red}.}
\label{tab:examples}
\end{table}
\begin{table}[t]
\resizebox{\columnwidth}{!}{
\begin{tabular}{@{}p{7cm}ccc@{}}
\toprule
\textbf{Sentence}                              & \textbf{Predicted}           & \textbf{GT} & \textbf{Confidence} \\ \midrule
Tin has the symbol Sn.                         & {\color[HTML]{32CB00} True}  & True        & 0.990                \\
Mercury is a liquid at room temperature and used in thermometers and some electrical switches. & {\color[HTML]{FE0000} False} & True & 0.113 \\
Fluorine is in the Alkaline earth metal group. & {\color[HTML]{32CB00} False} & False       & 0.012               \\
Tellurium has the atomic number of 82.         & {\color[HTML]{FE0000} True} & False        & 0.998               \\ \bottomrule
\end{tabular}}
\caption{Examples of predictions using SAPLMA and \llamatwo{}. In these examples, the probe classifier is trained on the original version of the True-False dataset~\cite{azaria2023llmlying} (excluding the \textit{element} topic) and tested on our refined version of the \textit{element} topic. Correct predictions are highlighted in {\color[HTML]{32CB00} green}, while incorrect ones are shown in {\color[HTML]{FE0000} red}.}
\label{tab:examples_mitch}
\end{table}


\begin{thebibliography}{34}
\providecommand{\natexlab}[1]{#1}

\bibitem[{Alain(2016)}]{alain2016understanding}
Guillaume Alain. 2016.
\newblock Understanding intermediate layers using linear classifier probes.
\newblock \emph{arXiv preprint arXiv:1610.01644}.

\bibitem[{Anelli et~al.(2022)Anelli, Biancofiore, Bellis, Noia, and Sciascio}]{DBLP:conf/cikm/AnelliBBNS22}
{Vito Walter} Anelli, Giovanni~Maria Biancofiore, Alessandro~De Bellis, Tommaso~Di Noia, and Eugenio~Di Sciascio. 2022.
\newblock \href {https://doi.org/10.1145/3511808.3557617} {Interpretability of {BERT} latent space through knowledge graphs}.
\newblock In \emph{Proceedings of the 31st {ACM} International Conference on Information {\&} Knowledge Management, Atlanta, GA, USA, October 17-21, 2022}, pages 3806--3810. {ACM}.

\bibitem[{Azaria and Mitchell(2023)}]{azaria2023llmlying}
Amos Azaria and Tom~M. Mitchell. 2023.
\newblock The internal state of an {LLM} knows when it's lying.
\newblock In \emph{{EMNLP} (Findings)}, pages 967--976. Association for Computational Linguistics.

\bibitem[{Biancofiore et~al.(2025)Biancofiore, Di~Palma, Pomo, Narducci, and Di~Noia}]{biancofiore2025conversational}
Giovanni~Maria Biancofiore, Dario Di~Palma, Claudio Pomo, Fedelucio Narducci, and Tommaso Di~Noia. 2025.
\newblock Conversational user interfaces and agents.
\newblock In \emph{Human-Centered AI: An Illustrated Scientific Quest}, pages 399--438. Springer.

\bibitem[{B{\"{u}}rger et~al.(2024)B{\"{u}}rger, Hamprecht, and Nadler}]{DBLP:conf/nips/BurgerHN24}
Lennart B{\"{u}}rger, Fred~A. Hamprecht, and Boaz Nadler. 2024.
\newblock Truth is universal: Robust detection of lies in llms.
\newblock In \emph{NeurIPS}.

\bibitem[{Calderon et~al.(2025)Calderon, Reichart, and Dror}]{DBLP:journals/corr/abs-2501-10970}
Nitay Calderon, Roi Reichart, and Rotem Dror. 2025.
\newblock The alternative annotator test for llm-as-a-judge: How to statistically justify replacing human annotators with llms.
\newblock \emph{CoRR}, abs/2501.10970.

\bibitem[{Chen et~al.(2024)Chen, Liu, Chen, Gu, Wu, Tao, Fu, and Ye}]{chen2024inside}
Chao Chen, Kai Liu, Ze~Chen, Yi~Gu, Yue Wu, Mingyuan Tao, Zhihang Fu, and Jieping Ye. 2024.
\newblock Inside: Llms' internal states retain the power of hallucination detection.
\newblock \emph{arXiv preprint arXiv:2402.03744}.

\bibitem[{Conneau et~al.(2018)Conneau, Kruszewski, Lample, Barrault, and Baroni}]{conneau2018you}
Alexis Conneau, German Kruszewski, Guillaume Lample, Lo{\"\i}c Barrault, and Marco Baroni. 2018.
\newblock What you can cram into a single vector: Probing sentence embeddings for linguistic properties.
\newblock \emph{arXiv preprint arXiv:1805.01070}.

\bibitem[{Dahl et~al.(2024)Dahl, Magesh, Suzgun, and Ho}]{Dahl2024}
Matthew Dahl, Varun Magesh, Mirac Suzgun, and Daniel~E. Ho. 2024.
\newblock Large legal fictions: Profiling legal hallucinations in large language models.
\newblock \emph{CoRR}, abs/2401.01301.

\bibitem[{{De Bellis} et~al.(2024){De Bellis}, Anelli, Noia, and Sciascio}]{DBLP:conf/semweb/BellisANS24}
Alessandro {De Bellis}, Vito~Walter Anelli, Tommaso~Di Noia, and Eugenio~Di Sciascio. 2024.
\newblock {PRONTO:} prompt-based detection of semantic containment patterns in {MLM}s.
\newblock In \emph{{ISWC} {(2)}}, volume 15232 of \emph{Lecture Notes in Computer Science}, pages 227--246. Springer.

\bibitem[{{Di Palma}(2023)}]{DBLP:conf/recsys/Palma23}
Dario {Di Palma}. 2023.
\newblock Retrieval-augmented recommender system: Enhancing recommender systems with large language models.
\newblock In \emph{RecSys}, pages 1369--1373. {ACM}.

\bibitem[{Di~Palma et~al.(2025)Di~Palma, Merra, Sfilio, Anelli, Narducci, and Di~Noia}]{di2025llms}
Dario Di~Palma, Felice~Antonio Merra, Maurizio Sfilio, Vito~Walter Anelli, Fedelucio Narducci, and Tommaso Di~Noia. 2025.
\newblock \href {https://doi.org/10.1145/3726302.3730178} {Do llms memorize recommendation datasets? a preliminary study on movielens-1m}.
\newblock In \emph{Proceedings of the 48th International {ACM} {SIGIR} Conference on Research and Development in Information Retrieval, {SIGIR} 2025, Padua, Italy July 13-18, 2025}. {ACM}.

\bibitem[{Duan et~al.(2024)Duan, Suri, Mireshghallah, Min, Shi, Zettlemoyer, Tsvetkov, Choi, Evans, and Hajishirzi}]{DBLP:journals/corr/abs-2402-07841}
Michael Duan, Anshuman Suri, Niloofar Mireshghallah, Sewon Min, Weijia Shi, Luke Zettlemoyer, Yulia Tsvetkov, Yejin Choi, David Evans, and Hannaneh Hajishirzi. 2024.
\newblock Do membership inference attacks work on large language models?
\newblock \emph{CoRR}, abs/2402.07841.

\bibitem[{Feng et~al.(2024)Feng, Shi, Wang, Ding, Balachandran, and Tsvetkov}]{feng-etal-2024-dont}
Shangbin Feng, Weijia Shi, Yike Wang, Wenxuan Ding, Vidhisha Balachandran, and Yulia Tsvetkov. 2024.
\newblock \href {https://doi.org/10.18653/v1/2024.acl-long.786} {Don`t hallucinate, abstain: Identifying {LLM} knowledge gaps via multi-{LLM} collaboration}.
\newblock In \emph{Proceedings of the 62nd Annual Meeting of the Association for Computational Linguistics (Volume 1: Long Papers)}, pages 14664--14690, Bangkok, Thailand. Association for Computational Linguistics.

\bibitem[{Gekhman et~al.(2025)Gekhman, Ben{-}David, Orgad, Ofek, Belinkov, Szpektor, Herzig, and Reichart}]{gekham}
Zorik Gekhman, Eyal Ben{-}David, Hadas Orgad, Eran Ofek, Yonatan Belinkov, Idan Szpektor, Jonathan Herzig, and Roi Reichart. 2025.
\newblock Inside-out: Hidden factual knowledge in llms.
\newblock \emph{CoRR}, abs/2503.15299.

\bibitem[{Gu et~al.(2024)Gu, Jiang, Shi, Tan, Zhai, Xu, Li, Shen, Ma, Liu, Wang, and Guo}]{DBLP:journals/corr/abs-2411-15594}
Jiawei Gu, Xuhui Jiang, Zhichao Shi, Hexiang Tan, Xuehao Zhai, Chengjin Xu, Wei Li, Yinghan Shen, Shengjie Ma, Honghao Liu, Yuanzhuo Wang, and Jian Guo. 2024.
\newblock A survey on llm-as-a-judge.
\newblock \emph{CoRR}, abs/2411.15594.

\bibitem[{Holtzman et~al.(2020)Holtzman, Buys, Du, Forbes, and Choi}]{DBLP:conf/iclr/HoltzmanBDFC20}
Ari Holtzman, Jan Buys, Li~Du, Maxwell Forbes, and Yejin Choi. 2020.
\newblock \href {https://openreview.net/forum?id=rygGQyrFvH} {The curious case of neural text degeneration}.
\newblock In \emph{8th International Conference on Learning Representations, {ICLR} 2020, Addis Ababa, Ethiopia, April 26-30, 2020}. OpenReview.net.

\bibitem[{Huang et~al.(2024)Huang, Yu, Ma, Zhong, Feng, Wang, Chen, Peng, Feng, Qin, and Liu}]{Huang2024-vq}
Lei Huang, Weijiang Yu, Weitao Ma, Weihong Zhong, Zhangyin Feng, Haotian Wang, Qianglong Chen, Weihua Peng, Xiaocheng Feng, Bing Qin, and Ting Liu. 2024.
\newblock A survey on hallucination in large language models: Principles, taxonomy, challenges, and open questions.
\newblock \emph{ACM Trans. Inf. Syst.}

\bibitem[{Ji et~al.(2023)Ji, Yu, Xu, Lee, Ishii, and Fung}]{ji-etal-2023-towards}
Ziwei Ji, Tiezheng Yu, Yan Xu, Nayeon Lee, Etsuko Ishii, and Pascale Fung. 2023.
\newblock \href {https://doi.org/10.18653/v1/2023.findings-emnlp.123} {Towards mitigating {LLM} hallucination via self reflection}.
\newblock In \emph{Findings of the Association for Computational Linguistics: EMNLP 2023}, pages 1827--1843, Singapore. Association for Computational Linguistics.

\bibitem[{Joshi et~al.(2017)Joshi, Choi, Weld, and Zettlemoyer}]{triviaqa}
Mandar Joshi, Eunsol Choi, Daniel~S. Weld, and Luke Zettlemoyer. 2017.
\newblock Triviaqa: {A} large scale distantly supervised challenge dataset for reading comprehension.
\newblock In \emph{{ACL} {(1)}}, pages 1601--1611. Association for Computational Linguistics.

\bibitem[{Kadavath et~al.(2022)Kadavath, Conerly, Askell, Henighan, Drain, Perez, Schiefer, Hatfield{-}Dodds, DasSarma, Tran{-}Johnson, Johnston, Showk, Jones, Elhage, Hume, Chen, Bai, Bowman, Fort, Ganguli, Hernandez, Jacobson, Kernion, Kravec, Lovitt, Ndousse, Olsson, Ringer, Amodei, Brown, Clark, Joseph, Mann, McCandlish, Olah, and Kaplan}]{kadavath2022}
Saurav Kadavath, Tom Conerly, Amanda Askell, Tom Henighan, Dawn Drain, Ethan Perez, Nicholas Schiefer, Zac Hatfield{-}Dodds, Nova DasSarma, Eli Tran{-}Johnson, Scott Johnston, Sheer~El Showk, Andy Jones, Nelson Elhage, Tristan Hume, Anna Chen, Yuntao Bai, Sam Bowman, Stanislav Fort, Deep Ganguli, Danny Hernandez, Josh Jacobson, Jackson Kernion, Shauna Kravec, Liane Lovitt, Kamal Ndousse, Catherine Olsson, Sam Ringer, Dario Amodei, Tom Brown, Jack Clark, Nicholas Joseph, Ben Mann, Sam McCandlish, Chris Olah, and Jared Kaplan. 2022.
\newblock Language models (mostly) know what they know.
\newblock \emph{CoRR}, abs/2207.05221.

\bibitem[{Levinstein and Herrmann(2024)}]{levinstein2024still}
Benjamin~A Levinstein and Daniel~A Herrmann. 2024.
\newblock Still no lie detector for language models: Probing empirical and conceptual roadblocks.
\newblock \emph{Philosophical Studies}, pages 1--27.

\bibitem[{Lin et~al.(2022)Lin, Hilton, and Evans}]{truthful_citation}
Stephanie Lin, Jacob Hilton, and Owain Evans. 2022.
\newblock Truthfulqa: Measuring how models mimic human falsehoods.
\newblock In \emph{{ACL} {(1)}}, pages 3214--3252. Association for Computational Linguistics.

\bibitem[{Marks and Tegmark(2023)}]{DBLP:journals/corr/abs-2310-06824}
Samuel Marks and Max Tegmark. 2023.
\newblock The geometry of truth: Emergent linear structure in large language model representations of true/false datasets.
\newblock \emph{CoRR}, abs/2310.06824.

\bibitem[{Orgad et~al.(2024)Orgad, Toker, Gekhman, Reichart, Szpektor, Kotek, and Belinkov}]{orgad2024}
Hadas Orgad, Michael Toker, Zorik Gekhman, Roi Reichart, Idan Szpektor, Hadas Kotek, and Yonatan Belinkov. 2024.
\newblock Llms know more than they show: On the intrinsic representation of {LLM} hallucinations.
\newblock \emph{CoRR}, abs/2410.02707.

\bibitem[{Pham and Vo(2024)}]{medical}
Duy~Khoa Pham and Bao~Quoc Vo. 2024.
\newblock Towards reliable medical question answering: Techniques and challenges in mitigating hallucinations in language models.
\newblock \emph{CoRR}, abs/2408.13808.

\bibitem[{Rajpurkar et~al.(2018)Rajpurkar, Jia, and Liang}]{squad_citation}
Pranav Rajpurkar, Robin Jia, and Percy Liang. 2018.
\newblock Know what you don't know: Unanswerable questions for squad.
\newblock In \emph{{ACL} {(2)}}, pages 784--789. Association for Computational Linguistics.

\bibitem[{Tenney et~al.(2019)Tenney, Xia, Chen, Wang, Poliak, McCoy, Kim, Van~Durme, Bowman, Das et~al.}]{tenney2019you}
Ian Tenney, Patrick Xia, Berlin Chen, Alex Wang, Adam Poliak, R~Thomas McCoy, Najoung Kim, Benjamin Van~Durme, Samuel~R Bowman, Dipanjan Das, et~al. 2019.
\newblock What do you learn from context? probing for sentence structure in contextualized word representations.
\newblock \emph{arXiv preprint arXiv:1905.06316}.

\bibitem[{Touvron et~al.(2023)Touvron, Martin, Stone, Albert, Almahairi, and et~al.}]{DBLP:journals/corr/abs-2307-09288}
Hugo Touvron, Louis Martin, Kevin Stone, Peter Albert, Amjad Almahairi, and Yasmine~Babaei et~al. 2023.
\newblock Llama 2: Open foundation and fine-tuned chat models.
\newblock \emph{CoRR}, abs/2307.09288.

\bibitem[{Upadhyay et~al.(2023)Upadhyay, Ginsberg, and Callison{-}Burch}]{UpadhyayGC23}
Shriyash Upadhyay, Etan Ginsberg, and Chris Callison{-}Burch. 2023.
\newblock Improving mathematics tutoring with {A} code scratchpad.
\newblock In \emph{BEA@ACL}, pages 20--28. Association for Computational Linguistics.

\bibitem[{Wadden et~al.(2020)Wadden, Lin, Lo, Wang, van Zuylen, Cohan, and Hajishirzi}]{wadden-etal-2020-fact}
David Wadden, Shanchuan Lin, Kyle Lo, Lucy~Lu Wang, Madeleine van Zuylen, Arman Cohan, and Hannaneh Hajishirzi. 2020.
\newblock \href {https://doi.org/10.18653/v1/2020.emnlp-main.609} {Fact or fiction: Verifying scientific claims}.
\newblock In \emph{Proceedings of the 2020 Conference on Empirical Methods in Natural Language Processing (EMNLP)}, pages 7534--7550, Online. Association for Computational Linguistics.

\bibitem[{Zhang et~al.(2022)Zhang, Roller, Goyal, Artetxe, Chen, Chen, Dewan, Diab, Li, Lin et~al.}]{zhang2022opt}
Susan Zhang, Stephen Roller, Naman Goyal, Mikel Artetxe, Moya Chen, Shuohui Chen, Christopher Dewan, Mona Diab, Xian Li, Xi~Victoria Lin, et~al. 2022.
\newblock Opt: Open pre-trained transformer language models.
\newblock \emph{arXiv preprint arXiv:2205.01068}.

\bibitem[{Zhang et~al.(2024)Zhang, Yao, Zhang, Yun, Yu, Li, and Tang}]{zhang2024transferable}
Xiaokang Zhang, Zijun Yao, Jing Zhang, Kaifeng Yun, Jifan Yu, Juanzi Li, and Jie Tang. 2024.
\newblock Transferable and efficient non-factual content detection via probe training with offline consistency checking.
\newblock \emph{arXiv preprint arXiv:2404.06742}.

\bibitem[{Zhang et~al.(2023)Zhang, Li, Cui, Cai, Liu, Fu, Huang, Zhao, Zhang, Chen et~al.}]{zhang2023siren}
Yue Zhang, Yafu Li, Leyang Cui, Deng Cai, Lemao Liu, Tingchen Fu, Xinting Huang, Enbo Zhao, Yu~Zhang, Yulong Chen, et~al. 2023.
\newblock Siren's song in the ai ocean: a survey on hallucination in large language models.
\newblock \emph{arXiv preprint arXiv:2309.01219}.

\end{thebibliography}
\end{document}